\documentclass[
]{ceurart}

\usepackage[colorinlistoftodos,disable]{todonotes}

\usepackage[utf8]{inputenc}
\usepackage{graphicx}
\usepackage{url}
\usepackage{hyperref}
\usepackage{cleveref}
\usepackage{pgfplotstable} 
\usepackage{booktabs}
\usepackage{siunitx}
\usepackage{capt-of} 
\usepackage{outlines}
\usepackage{multirow}
\usepackage[inline]{enumitem}
\usepackage{subcaption}
\usepackage{nicefrac}

\usepackage{tikz}
\usepackage[dvipsnames]{xcolor}
\usepackage{tikz-network}

\usepackage[toc,page]{appendix} 

\usetikzlibrary{external}

\newcommand{\lotr}{\emph{The Lord of the Rings}}
\newcommand{\hobbit}{\emph{The Hobbit}}
\newcommand{\sil}{\emph{The Silmarillion}}
\newcommand{\leg}{\emph{Legendarium}}

\DeclareMathOperator{\ct}{count}
\DeclareMathOperator{\argmax}{argmax}

\begin{document}

\copyrightyear{2022}
\copyrightclause{Copyright for this paper by its authors.
  Use permitted under Creative Commons License Attribution 4.0
  International (CC BY 4.0).}

\conference{CHR 2022: Computational Humanities Research Conference, December 12 -- 14, 2022, Antwerp, Belgium}

\title{One Graph to Rule them All: Using NLP and Graph Neural Networks to analyse Tolkien's Legendarium}

\author[1]{Vincenzo Perri}[%
email=perri@ifi.uzh.ch,
]
\cormark[1]
\fnmark[1]
\address[1]{Data Analytics Group, Department of Informatics(IfI), Universität Zürich, CH-8050 Zürich, Switzerland}

\author[2]{Lisi Qarkaxhija}[%
email=lisi.qarkaxhija@uni-wuerzburg.de,
]
\cormark[1]
\fnmark[1]
\address[2]{Chair of Informatics XV - Machine Learning for Complex Networks, Center for Artificial Intelligence and Data Science (CAIDAS), Julius-Maximilians-Universität Würzburg, D-97074 Würzburg, Germany}

\author[3]{Albin Zehe}[%
email=zehe@informatik.uni-wuerzburg.de,
]
\cormark[1]
\fnmark[1]
\address[3]{Chair of Informatics X - Data Science, Center for Artificial Intelligence and Data Science (CAIDAS), Julius-Maximilians-Universität Würzburg, D-97074 Würzburg, Germany}

\author[3]{Andreas Hotho}[%
email=hotho@informatik.uni-wuerzburg.de,
]
\cormark[1]
\address[3]{Chair of Informatics X - Data Science, Center for Artificial Intelligence and Data Science (CAIDAS), Julius-Maximilians-Universität Würzburg, D-97074 Würzburg, Germany}

\author[2,1]{Ingo Scholtes}[%
email=ingo.scholtes@uni-wuerzburg.de,
]

\cortext[1]{Corresponding author.}
\fntext[2]{These authors contributed equally.}

\begin{abstract}
    Natural Language Processing and Machine Learning have considerably advanced Computational Literary Studies. Similarly, the construction of co-occurrence networks of literary characters, and their analysis using methods from social network analysis and network science, have provided insights into the micro- and macro-level structure of literary texts. Combining these perspectives, in this work we study character networks extracted from a text corpus of J.R.R. Tolkien's Legendarium. We show that this perspective helps us to analyse and visualise the narrative style that characterises Tolkien's works. Addressing character classification, embedding and co-occurrence prediction, we further investigate the advantages of state-of-the-art Graph Neural Networks over a popular word embedding method. Our results highlight the large potential of graph learning in Computational Literary Studies.
\end{abstract}


\maketitle

\section{Motivation and Background}
\label{sec:intro}

Computational Literary Studies (CLS) have recently taken advantage of the latest developments in Natural Language Processing (NLP), with deep learning techniques bringing major improvements for tasks relevant to literature analysis:
Examples include named entity recognition \cite{li2020survey}, anaphora and coreference resolution \cite{sukthanker2020anaphora}, sentiment analysis \cite{feldman2013techniques}, scene detection \cite{zehe2021detecting} or genre classification \cite{worsham2018genre}.
Apart from NLP, machine learning has recently shown great potential in data with complex \emph{relational} structure that can be represented as \emph{graph} or \emph{network} $G=(V,E)$, consisting of nodes $u, v, ... \in V$ and links $(u,v) \in E$.
In CLS, this abstraction is frequently used to study \emph{character networks}, i.e. graphs where nodes represent literary characters and links represent relationships such as, e.g., their co-occurrence in a sentence or scene, or dialogue interactions \cite{labatut2019extraction}.
Building on this abstraction, several works in CLS used (social) network analysis to study the narrative structure of literary works \cite{moretti2011network,trilcke2013social}:
Considering 19th century novels, \citet{elson2010extracting} analysed macroscopic properties of conversation networks, i.e. fictional characters engaging in dialogue.
\citet{beveridge2016network} applied centrality measures and community detection to a network of characters that occur within close proximity in the text of \emph{A Storm of Swords}, the third novel in George R.R. Martin's series \emph{Song of Ice of Fire}.
Using the same network extraction, \citet{bonato2016mining} studied statistical properties of character networks in three popular novels.
Several authors applied centrality measures to identify important characters, e.g. in works by Shakespeare \cite{yavuz2020analyses}, \emph{Alice in Wonderland} \cite{agarwal2012social}, or the first novel of the \emph{Harry Potter} series \cite{sparavigna2013social}.
In a recent work, \citet{agarwal2021genre} used character networks to facilitate the automated classification of literary genres. 
Using J.R.R. Tolkien's \lotr{}, which we also consider in our manuscript, \citet{ribeiro2016complex} applied social network analysis to its network of characters.
\citet{li2019complex} studied small-world and scale-free properties of character co-occurrence networks in movie scripts, among them the movie adaptation of \lotr{}.

Existing studies of character networks mainly used methods from (social) network analysis to gain insights into the narrative structure of literary texts.
At the same time, recent advances in \emph{geometric deep learning} \cite{bronstein2017geometric} and \emph{Graph Neural Networks} (GNNs) provide new ways to apply deep learning to graph-structured data, which creates interesting opportunities for the study of character networks.
A major advantage of GNNs over other network analysis or machine learning techniques is their ability to leverage relational patterns in the topology of a graph, while at the same time incorporating additional node or link features.
This facilitates unsupervised and (semi-)supervised learning tasks at the node, link, or graph level.
Examples include graph representation learning \cite{perozzi2014deepwalk, grover2016node2vec}, community detection \cite{bruna2017community}, node and link classification \cite{Kipf:2016tc, velikovi2017graph}, link prediction \cite{zhang2018link}, or graph classification \cite{10.5555/3504035.3504579}.
To the best of our knowledge, no works have combined recent advances in (i) natural language processing, e.g. to recognize entities, resolve coreferences, extract meaningful character networks, or generate word embeddings, and (ii) graph neural networks, e.g. to compute latent space representations of fictional characters or address node- and link-level learning tasks.
In CLS, the combination of these paradigms can help with several tasks, e.g. (semi-)supervised character classification, semantic analyses of character relationships, comparisons of character constellations in different works, or automated genre classification.
Addressing this gap, we combine NLP, network analysis and graph learning to analyse J.R.R. Tolkien's Legendarium, namely \sil{}, \hobbit{}, and the three volumes of \lotr{}.
Our contributions are:
\begin{itemize}[noitemsep,topsep=0pt]
    \item We apply entity recognition and coreference resolution to detect and disambiguate characters in Tolkien's Legendarium. We report key statistics like character mentions via pronouns, nominal mentions or explicit references and compare them across different works in the Legendarium. We use sequential character mentions to generate narrative charts of the five considered works, which highlight Tolkien's \emph{interlaced narrative style}.
    \item We extract \emph{character networks} for each work in our corpus, i.e. graphs $G=(V,E)$ where nodes $v \in V$ capture characters in the Legendarium, while undirected edges $(v,w)$ represent the co-occurrence of two characters in the same sentence.
    Apart from generating character networks for the five works in our corpus, we generate a \emph{single} network that captures all characters in the works that constitute Tolkien's Legendarium. We use this to perform a macroscopic, graph-based characterisation of Tolkien's works.
    \item  We address the question to what extent graph learning techniques can leverage the topology of automatically extracted character networks, and which advantages they provide over word embeddings that just consider word-context pairs.
    To this end, we evaluate the performance of different methods for the (i) latent space representation of characters, (ii) (semi-)supervised classification of characters, and (iii) prediction of character co-occurrences.
    The results confirm that the inclusion of topological information from character networks considerably improves the performance of these tasks.
\end{itemize}

Analysing a single graph representation of multiple literary works unfolding in the same fictional universe, our work demonstrates the potential of graph neural networks for computational literary studies.
To facilitate the use of our data and methods in the (digital) study of Tolkien's works\footnote{\url{https://digitaltolkien.com/}}, we make the results of our entity recognition, coreference resolution and character network extraction available.
To foster the application of our methods to other corpora, we also provide a set of \text{jupyter} notebooks that reproduce our findings.\footnote{Will be released upon acceptance.}

\section{Text Corpus and Data Processing}
\label{sec:data_processing}

We consider the English full text of \lotr{} (consisting of the volumes \emph{The Fellowship of the Ring}, \emph{The Two Towers} and \emph{The Return of the King}, each split into two books), \hobbit{} and \sil{}.
We used the python-based NLP pipeline \texttt{BookNLP}\footnote{\url{https://github.com/booknlp/booknlp}.} to extract linguistic and literary features.
\texttt{BookNLP} uses the NLP service \texttt{spaCy} \cite{honnibal2020spacy} to perform tokenisation, that is, splitting text into a list of words and special characters (like punctuation) and to split text into sentences.
This first step in the NLP pipeline is the basis for all further processing steps.

\paragraph{Entity Recognition and Coreference Resolution}
Entity recognition refers to the task of detecting all references to \emph{entities} (e.g., characters, location) in a text corpus.
These references can either be explicitly named references (e.g., ``Bilbo Baggins'', ``Smaug''), noun phrases (e.g., ``the hobbit'', ``the dragon``) or pronouns (e.g., ``she'', ``they'').
\texttt{BookNLP} uses an entity annotation model that has been trained on a large annotated data set \cite{bamman2019annotated} to identify named entities, noun phrases as well as pronoun references.
After these references have been detected, in a next step coreference resolution can be applied, which is a very hard task in general \cite{Sukthanker2020} and is especially hard in the context of literary texts due to the high variation of references used and the very long texts \cite{schroeder2021neural,Krug2020}.
Confirming this view, our initial analyses revealed that the performance of \texttt{BookNLP}'s coreference resolution, which was trained on a data set of annotated coreferences \cite{bamman2019annotated2} was not satisfactory when applying it to our corpus.
We thus decided to focus on named references, and resolve these using a set of simple manually-created disambiguation rules (e.g., ``Sam'' $\rightarrow$ ``Sam Gamgee'', ``Peregrin'' $\rightarrow$ ``Pippin'').\footnote{The full list of disambiguation rules can be found in our repository.}
Although this approach may yield a low recall (i.e. there are many unidentified coreferences since pronouns and noun phrases are not considered), we find that this coreference resolution yields high precision (i.e. almost all resolved coreferences that we inspected manually were correct).
We found this approach preferable over a ``full'' coreference resolution for two reasons: 
First, considering our focus on character networks, a coreference resolution with high recall but lower precision would give rise to many spurious character co-occurrences that would harm our analyses of graph learning techniques. 
Second, our corpus of Tolkien's Legendarium is special in the sense that it has a large number of named references, which give rise to rich character networks despite limiting our view to named references.

\paragraph{Extraction of Character Co-Occurrences}
After finding references to characters, we next extract co-occurrences of pairs of characters that can be used to build character networks.
While the co-occurrence of characters does not necessarily imply an interaction between them, due to its simplicity it is a frequently used approach to construct character networks in CLS \cite{beveridge2016network,labatut2019extraction,bonato2016mining}.
After evaluating different strategies (cf. \Cref{sec:app:cooc}), we decided to extract each co-occurrence of two characters in the same sentence.

\paragraph{Descriptive Statistics of Text Corpus}
In \Cref{tab:dataset_stats}, we provide key summary statistics that characterize the different works in our corpus.
Apart from differences in terms of tokens, we find striking differences between character mentions in the five texts:
\sil{} uses considerably fewer pronoun mentions than the other four works, while using more explicit references by name.
We attribute this to the compressed writing style of \sil{}, which rather resembles a fictional historical record that lists character names and locations, and gives a chronology of events compared to the more conventional prose style of \hobbit{} and \lotr{}.

\begin{table}[!ht]
    \pgfkeys{/pgf/number format/.cd,fixed,precision=2}
    \begin{filecontents*}{data/stats.csv}
nan,Hobbit,TLotR Vol. I,TLotR Vol. II,TLotR Vol. III,Silmarillion
\#tokens,113235.0,224156.0,189539.0,163191.0,147833.0
\#character mentions,15477.0,31770.0,27576.0,24766.0,26468.0
\% Pronouns,56.18659947018156,54.107648725212464,56.2735712213519,51.647419849794076,37.475442043222
\% Nominal mentions,27.699166505136652,22.760465848284543,22.30200174064404,25.934749253008153,29.14840562188303
\% Explicit references,16.114234024681785,23.13188542650299,21.424427038004062,22.41783089719777,33.37615233489497
Character Co-occurrences,431,886,615,844,1674
\end{filecontents*}
\pgfplotstabletypeset[
        col sep=comma,
        row sep=newline,
        every column/.style={column type={r}},
        display columns/0/.style={column type={l}, column name={}, string type}
        ]{data/stats.csv}
    \caption{Summary statistics of the text corpus used in our analysis.}
    \label{tab:dataset_stats}
\end{table}

\begin{figure}[!b]
    \centering   
    \begin{subfigure}[c]{0.3\textwidth}
        \includegraphics[width=\textwidth]{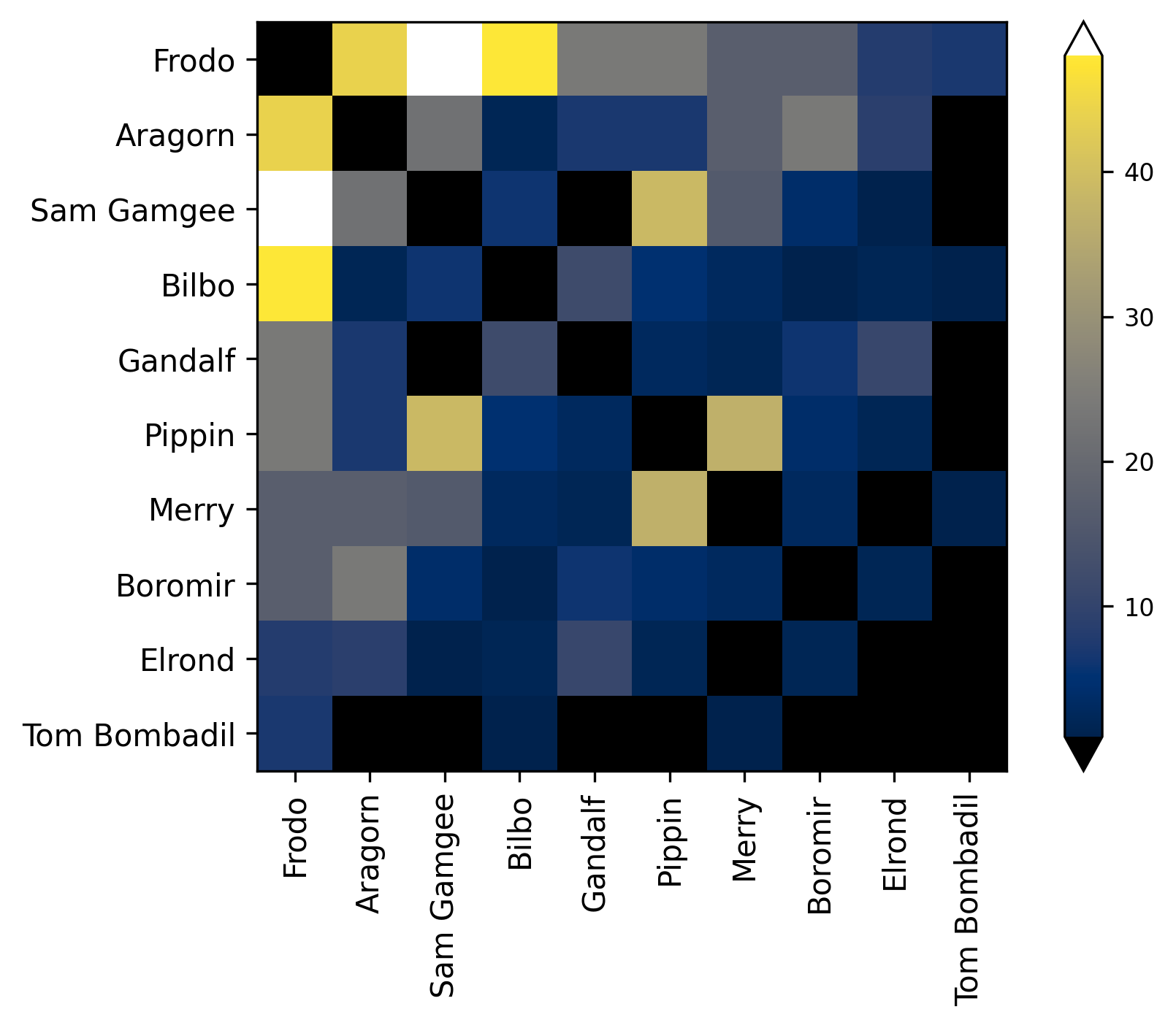}
        \subcaption{TLoTR, Vol. I\label{fig:interactions:lotr1}}
    \end{subfigure}
    \quad
    \begin{subfigure}[c]{0.3\textwidth}
        \includegraphics[width=\textwidth]{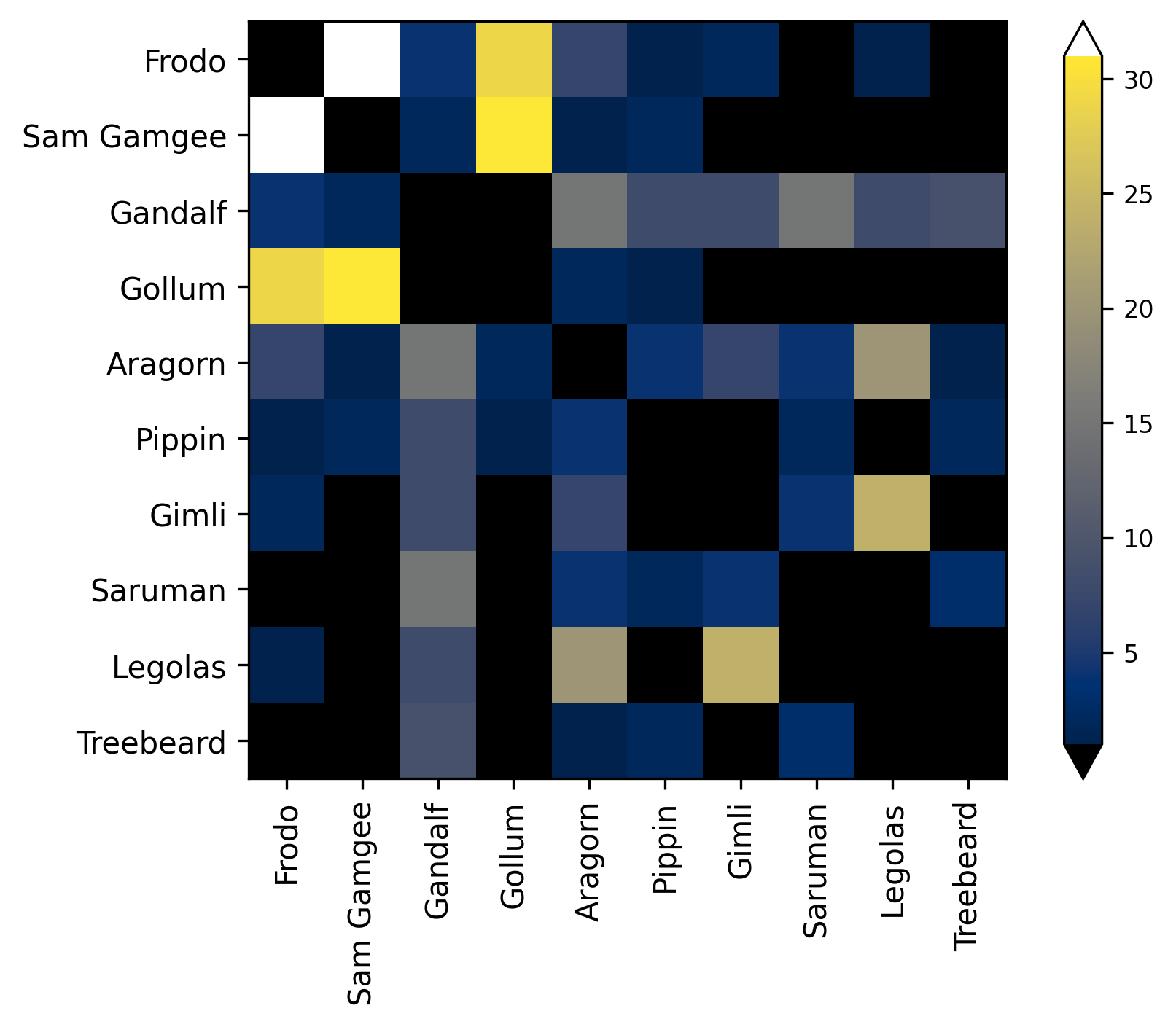}
        \subcaption{TLoTR, Vol. II\label{fig:interactions:lotr2}}
    \end{subfigure}
    \quad
    \begin{subfigure}[c]{0.3\textwidth}
        \includegraphics[width=\textwidth]{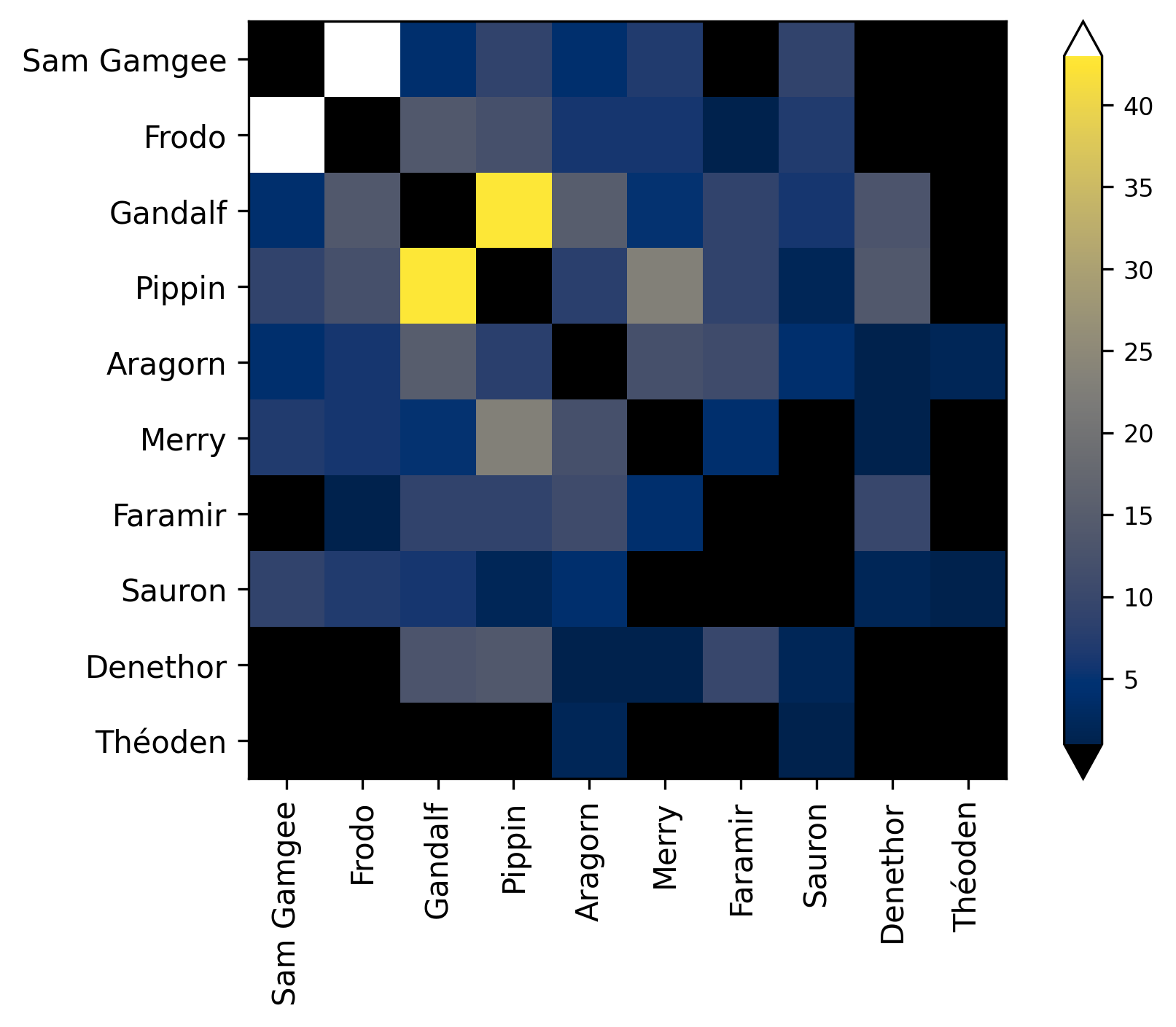}
        \subcaption{TLoTR, Vol. III\label{fig:interactions:lotr3}}
    \end{subfigure}
    \begin{subfigure}[c]{0.3\textwidth}
        \includegraphics[width=\textwidth]{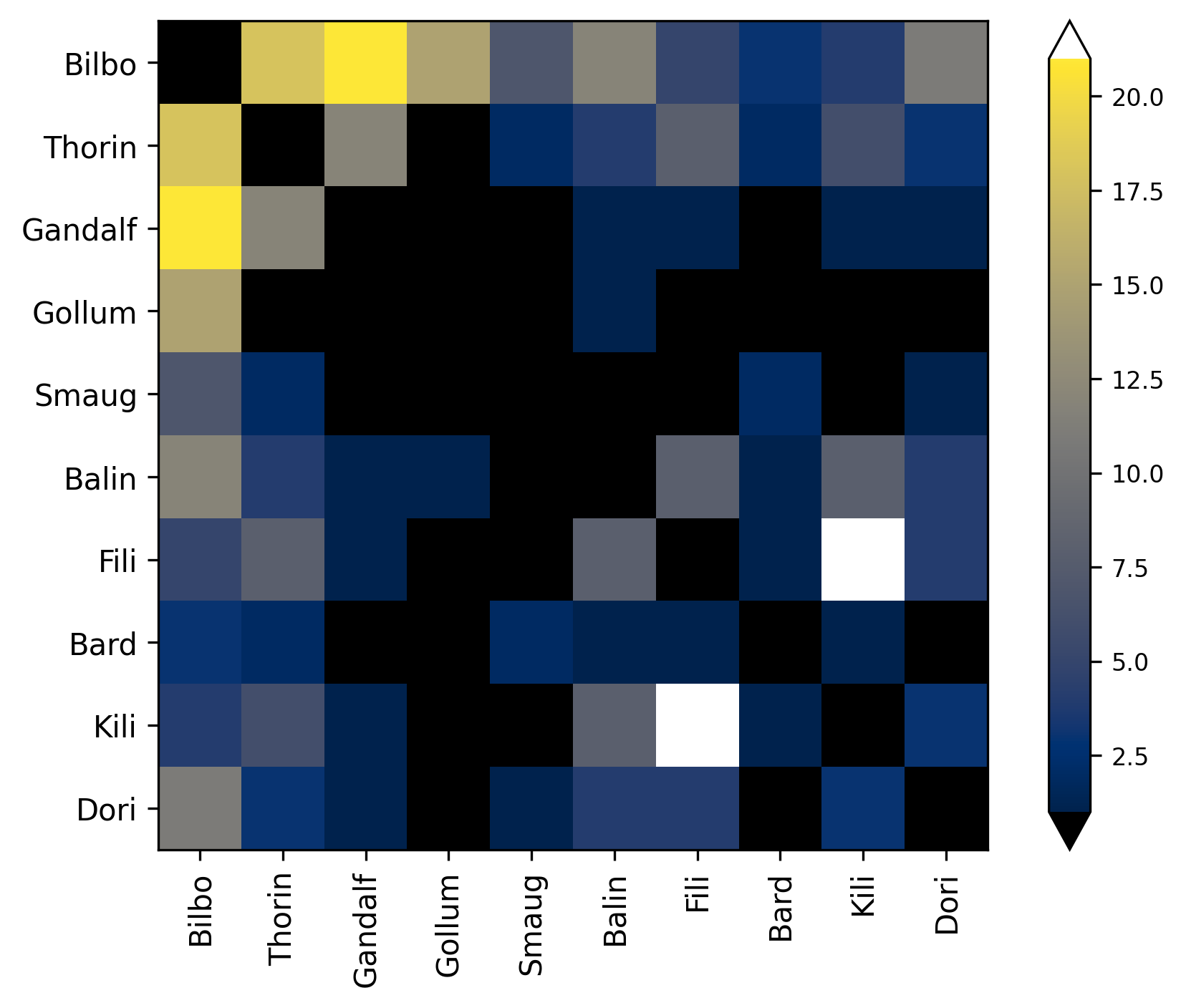}
        \subcaption{\hobbit{}\label{fig:interactions:hobbit}}
    \end{subfigure}
    \quad
    \begin{subfigure}[c]{0.3\textwidth}
        \includegraphics[width=\textwidth]{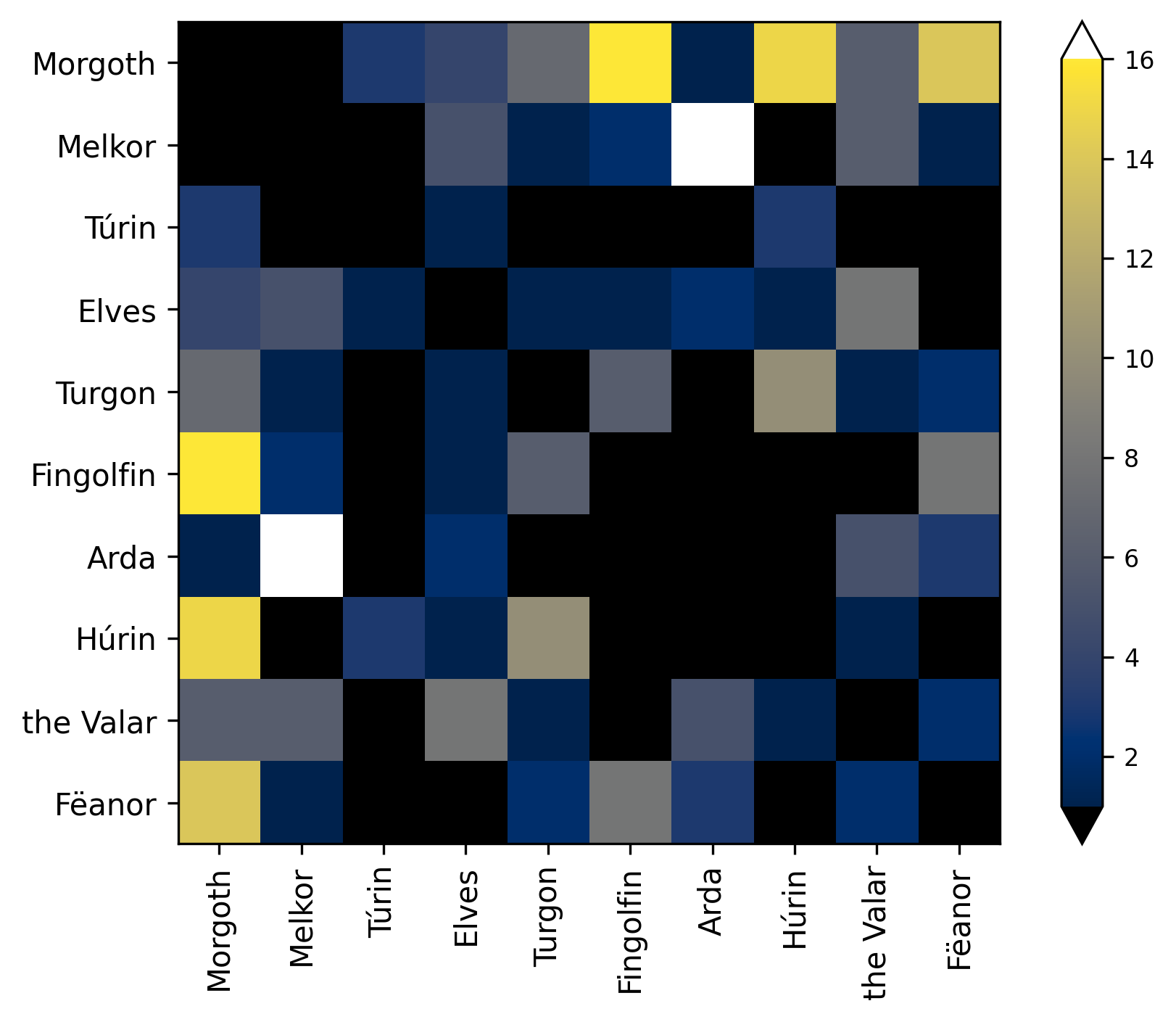}
        \subcaption{\sil{}\label{fig:interactions:sil}}
    \end{subfigure}
    \quad
    \begin{subfigure}[c]{0.3\textwidth}
        \includegraphics[width=\textwidth]{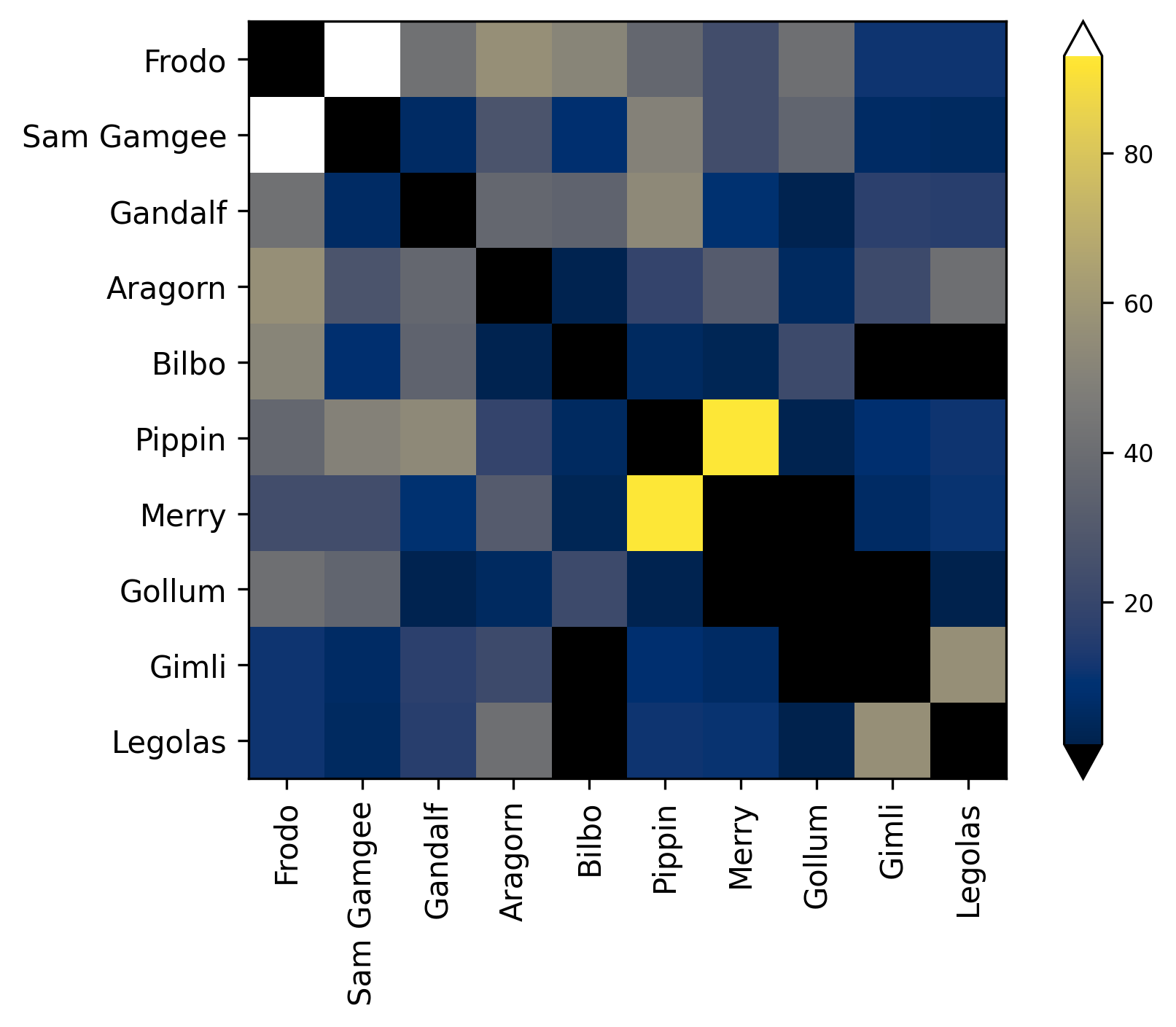}
        \subcaption{\leg{}\label{fig:interactions:legendarium}}
    \end{subfigure}
    \caption{Number of co-occurrences between the ten most frequently mentioned characters in the five works of our corpus (a) -- (e) and the whole Legendarium (f).\label{fig:interactions}}
\end{figure}

We calculate the number of character co-occurrences for each work in our corpus, which we indicate in \Cref{tab:dataset_stats} and visualise in \Cref{fig:interactions:lotr1} -- \Cref{fig:interactions:legendarium}.
The overall number of co-occurrences is influenced by the number of explicit references (cf. \Cref{tab:dataset_stats}), since we only extract co-occurrences if both characters are mentioned by name in the same sentence.

\paragraph{Character-based Visualization of Narrative Structure}
\label{sec:narrative}

\begin{figure}[!htb]
    \centering
    \includegraphics{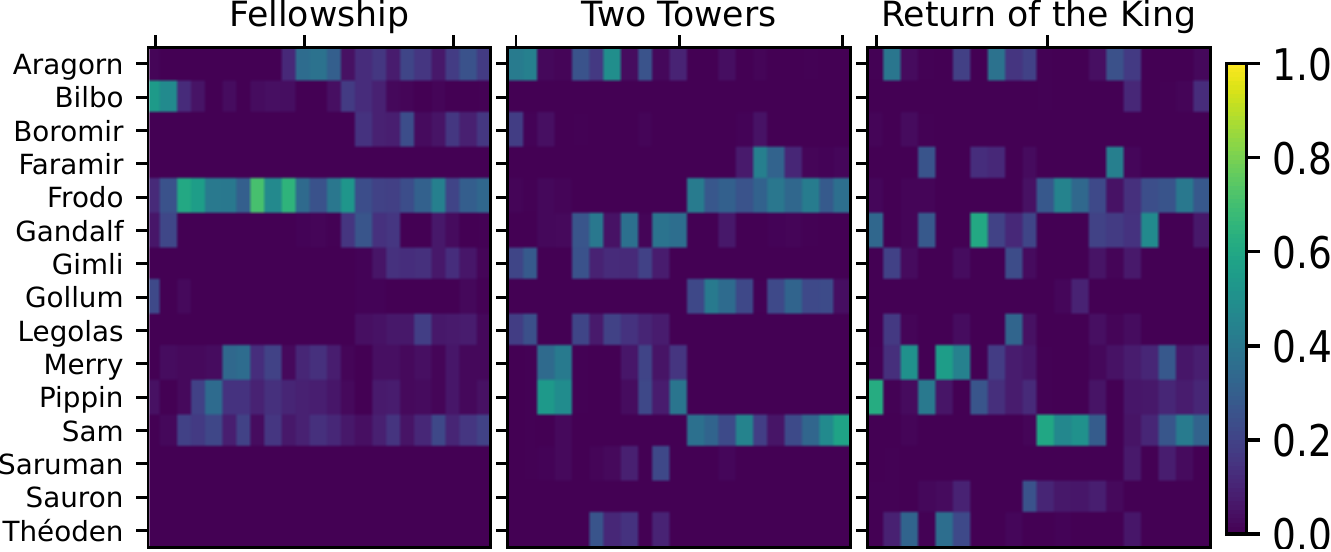}    
    \caption{Narrative charts for the three volumes of \lotr{}. Rows indicate key characters, while columns represent chapters. Colours capture the relative frequency of character mentions relative to other key characters captured in the charts. In Vol. I (left panel) Frodo and the Hobbits maintain a central role across the whole volume. For Vol. II and III (middle and right panel) clear transitions in the narrative structure are visible, which are due to an \emph{interlacing narrative style}. \label{fig:narrative_chart}}
   \end{figure}

We finally use the character mentions in conjunction with the chapter in which they occurred to automatically produce a narrative chart for the three volumes of \lotr{}.
To avoid occurrences of characters that are only mentioned while not being present, we excluded mentions of characters within dialogues, as detected by \texttt{BookNLP}.
The narrative charts for the three volumes of \lotr{} are shown in the three panels of \cref{fig:narrative_chart}.
The columns within each panel represent chapters.
To ease the presentation, we focus on a selected set of main characters shown in the rows. 
The color of each row-column cell represents the fraction of the number of times a characters is mentioned in a given chapter, relative to the number of mentions of other key characters presented in the narrative chart.

While it is easy to generate those charts, we can use them to identify major plot lines and reveal a narrative structure that is characteristic for \lotr{}:
In volume I, we see that Frodo and the Hobbits maintain a chief role throughout book I and II, i.e. both parts of volume I of \lotr{}.
A shift is visible for the second half (book II), which coincides with the Hobbits' arrival in Rivendell.
In volume II, we see a clear transition in narrative structure that is due to Tolkien's adoption of an \emph{interlacing} narrative style \cite{west1975interlace}.
The first half of volume II (book III) focuses on the main characters Aragorn, Legolas, and Gimli, and their attempt to rescue Merry and Pippin from the Uruk-Hai. 
Leaping back in time, the second half of volume III (book IV) focuses on the journey of Frodo, Sam, Gollum and their encounter with Faramir, which coincides with a brief absence of Gollum as he hides from the rangers.
Following the original titles suggested for the six books that constitute the three volumes of \lotr{} \cite{shippey2014road}, these interlaces can be called \emph{The Treason of Isengard} and \emph{The Journey of the Ring-bearers}.
In volume III, we see a similar but less marked separation between two interlaced plot lines:
The first half (book V) focuses on the War of the Ring in Gondor, while book VI continues the story of Frodo's and Sam's journey to Mount Doom, followed by the Scouring of the Shire, which explains Merry's and Saruman's reappearance in the last part.
The original titles referring to those interlaces are \emph{The War of the Ring} and \emph{The End of the Third Age}.
The non-linear narrative style expressed in the narrative charts in \Cref{fig:narrative_chart} is common in early medieval literature \cite{leyerle1967interlace}. 
Tolkien likely adopted it due to his familiarity with medieval literature, to increase suspense and to achieve a sense of historicity \cite{auger2008lord}.
Since these charts capture the narrative structure that we would expect, we can assume that our character extraction, even though using only limited coreference resolution, captures the overall occurrences of key characters rather well.
In \Cref{sec:app:narrative} we include additional narrative charts for \hobbit{} and \sil{}.

\section{Analysis of Character Co-occurrence Networks}
\label{sec:network_analysis}

\begin{figure}[!htp]
\centering
    \includegraphics[width=\textwidth]{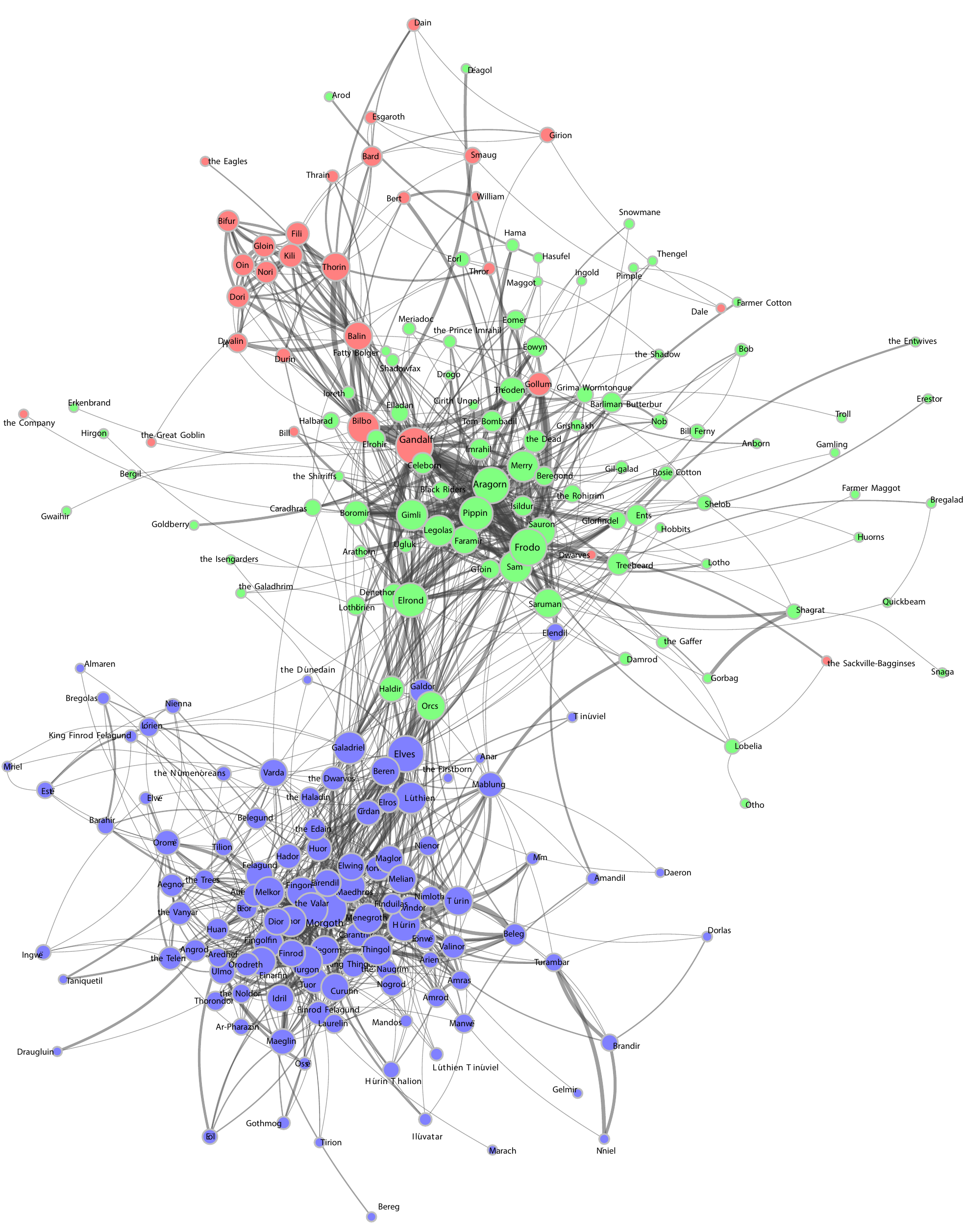}
    \caption{Force-directed visualisation of automatically extracted character network for Tolkien's Legendarium with $238$ nodes and $1233$ links. Node colours indicate the works in which a character occurs most frequently (red: \hobbit{}, green: \lotr{}, blue: \sil{}). Node sizes are logarithmically scaled based on node degrees, edge thickness are logarithmically scaled based on edge weights, i.e. number of character co-occurrences. Visualisation was created using the Open Source network analysis and visualisation library \texttt{pathpy} \cite{pathpy}. This character networks facilitates a macroscopic, graph-based analysis that considers intertextual aspects in J.R.R. Tolkien's Legendarium.\label{fig:main_vis}}
\end{figure}

Building on \Cref{sec:data_processing} we now use character co-occurrences to construct \emph{character networks} for J.R.R. Tolkien's Legendarium.
A \emph{graph} or \emph{network} is defined as tuple $G = (V, E)$, where $u, v, ... \in V$ represent nodes and $(u,v) \in E \subseteq V \times V$ is a set of directed or undirected links.
For our analysis, we model characters as nodes $V$ and their co-occurrences within a sentence as undirected links $E$, i.e. $(u,v) \in E \implies (v,u) \in E$ for all $u,v \in V$.
We further add link weights $w: E \rightarrow \mathbb{N}$, i.e. we assign a value $w((u,v))$ to each link $(u,v)$ that counts the co-occurrences of $u$ and $v$.
We adopt this definition to construct character networks for each of the four works in our corpus. 
A particularly interesting aspect of our corpus is that all works refer to a single Legendarium, with frequent cross-reference, and thus, overlaps in terms of characters.
We use this to construct a \emph{single} network of characters across all works, which we call \emph{Legendarium Graph}.
Except for a single disconnected pair of nodes that we removed, the resulting graph has a single connected component with $238$ nodes, which enables a macroscopic analysis of Tolkien's Legendarium.
\Cref{fig:main_vis} shows a visualisation of the character network that has been generated using the force-directed layout algorithm by \citet{fruchterman1991graph}.
This algorithm simulates attractive forces between connected nodes (and repulsive forces between all nodes) such that their positions in the stable state of the resulting dynamical system highlight patterns in the topology.

\begin{table}[!ht]
    \begin{tabular}{lrlrrrr}
\toprule
{} &  nodes &  edges &   density &  mean degree & diam. & avg. shortest path \\
\midrule
\hobbit{}             &     30 &    119 &  0.14 &     8.0 &  4    &        1.9 \\
\lotr{} Vol. I         &     65 &    217 &  0.052 &     6.68 &  5    &         2.31 \\
\lotr{} Vol. II         &     55 &    169 &  0.057 &     6.13 &  5    &          2.4 \\
\lotr{} Vol. III &     45 &    159 &  0.08 &     7.07 &    5    &     2.26 \\
\sil{}       &    136 &    803 &  0.044 &    11.81 &    5    &       2.41 \\
\midrule 
\emph{Legendarium} & 238 & 1233 & 0.023 & 10.48 & 6 & 2.84 \\ 
\bottomrule
\end{tabular}

    \caption{Key network metrics for the character networks in our corpus, where \emph{Legendarium} refers to the graph spanning \emph{all} works.}
    \label{table:standard_network_metrics}
\end{table}

\paragraph{Network Size, Density, and Mean Degree} 
Modelling character networks enables us to compute metrics from social network analysis, which we report in \Cref{table:standard_network_metrics}.
The first columns of \Cref{table:standard_network_metrics} show the number of nodes $n=|V|$ and links $m=|E|$ and the density $\rho = \frac{m}{n(n-1)}$ of each character network, where the latter captures which fraction of possible links actually occurred.
We find that \hobbit{} and \sil{} have the smallest and largest number of nodes respectively.
\hobbit{} has a higher link density, which is likely because the plot is focused on a small number of strongly interacting characters.
For \lotr{} we see that the number of nodes for Vol. III is smaller than for the previous two volumes, while the density is larger. 
This reflects the fact that the last volume is strongly focused on interactions between small groups of characters, e.g. Frodo, Sam and Gollum.
The degree of a node $v \in V$ is defined as the number of other nodes to which it is connected by links, i.e. $d(v) := |\{u: (u,v) \in E\}|$.
The fourth column in \Cref{table:standard_network_metrics} reports the mean node degree of characters, where larger mean degrees are associated with higher link density.

\begin{table}[!ht]
    \centering
    \begin{tabular}{lll}
\toprule
{} &   degree & betweennes \\
\midrule
1  &    Frodo &      Elves \\
2  &      Sam &      Frodo \\
3  &  Gandalf &    Gandalf \\
4  &  Aragorn &      Bilbo \\
5  &   Pippin &  Galadriel \\
6  &    Merry &    Aragorn \\
7  &  Morgoth &    Morgoth \\
8  &    Bilbo &     Elrond \\
9  &  Legolas &    Lúthien \\
10 &    Gimli &       Orcs \\
11 &   Elrond &        Sam \\
12 &   Gollum &    Saruman \\
13 &    Elves &     Pippin \\
14 &    Túrin &      Merry \\
15 &    Húrin &      Húrin \\
\bottomrule
\end{tabular}

    \caption{Ranking of characters in Tolkien's \emph{Legendarium} based on degree and betweenness centrality.\label{table:main_centralities}} 
\end{table}

\paragraph{Shortest paths, Diameter and Betweenness Centrality}
An important feature of (social) networks is the structure of shortest paths between nodes, which provide a topological notion of pair-wise distances \cite{newman2018networks}.
We calculate shortest paths between all pairs of nodes and report the average shortest path length across all character pairs.
We further calculate the diameter, i.e. the maximum shortest path length between any pair of nodes.
The results in \Cref{table:standard_network_metrics} show that, on the one hand, the average shortest path length is associated with the number of nodes (smallest average shortest path length for the smallest networks \hobbit{} and \lotr{} Vol. III).
On the other hand, it is influenced by mean degree and link density, which explains why \lotr{} Vol. III and \sil{} have similar values despite the latter having more than twice as many nodes.
The shortest paths between characters allows us to define a path-based notion of centrality.
The \emph{betweenness centrality} \cite{wasserman1994social} of node $v$ captures the number of shortest paths between any pair of nodes that pass through $v$, i.e. nodes have higher betweenness centrality if they are important for the ``communication'' between other nodes.
The second column of \cref{table:main_centralities} reports characters with the highest betweenness centrality for the single \emph{Legendarium} graph.
We note the high betweenness centrality  \emph{Galadriel}, who despite an ephemeral appearance in \lotr{} and absence in \hobbit{} is one of few characters that link the narrative across different ages in Tolkien's mythology.

\section{Application of Graph Neural Networks}
\label{sec:gnn}

We now turn our attention to the application of state-of-the-art graph learning techniques to the character network of Tolkien's Legendarium, which has been introduced and characterised in the previous section.
Our analysis is focused on our guiding research question outlined in \Cref{sec:intro}, i.e. what additional information we can draw from the topology of the character network, compared to an application of standard machine learning to a word embedding technique.
A major hurdle for the application of machine learning to character networks is that standard techniques like, e.g., logistic regression, support vector machines, or neural networks require features in a continuous vector space.
Their application to discrete objects like graphs typically requires a two-step procedure that consists of (i) a representation learning or embedding step to extract vectorial features of nodes and/or links, and (ii) a downstream application of machine learning to those features.
This approach is limited by the fact that graphs with complex topologies are fundamentally non-Euclidean objects \cite{bronstein2017geometric}, which limits our ability to find a generic vector space representation that is suitable for different learning tasks.

Addressing this limitations, recent works in the field of \emph{Geometric Deep Learning} \cite{bronstein2017geometric} and \emph{Graph Learning} have generalized deep learning to graph-structured data.
Among those works, Graph Neural Networks (GNNs) \cite{10.1109/72.572108,Gori2005ANM,10.1109/TNN.2008.2005605,Gallicchio} have developed into a particularly successful paradigm.
A major advantage of GNNs over other network analysis or machine learning techniques is their ability to capture both relational patterns in the topology of a graph as well as additional (vectorial) features of nodes or links.
A common concept of GNNs is their use of hidden layers with \emph{neural message passing}, i.e. nodes repeatedly exchange and update feature vectors with neighbouring nodes thus incorporating information from their neighbourhood.
The (hidden) features generated in this way can be used to address learning tasks by means of a perceptron model with a non-linear activation function.
The gradient-based optimization of GNNs can be thought of as an implicit way to generate a topology- and feature-aware latent space representation of a graph that facilitates node-, link- or graph-level learning tasks \cite{glrbook}.

Moving beyond the social network analysis techniques applied in \Cref{sec:network_analysis}, in the following we apply two state-of-the-art GNN architectures to the character network of Tolkien's Legendarium.
We use them to address three unsupervised and supervised learning tasks: (i) representation learning, i.e. finding a meaningful latent space embedding of characters, (ii) (semi-)supervised node classification, i.e. assigning characters to different works in the Legendarium, and (iii) link prediction, i.e. using a subset of links in the graph to predict missing links in a holdout set.

\paragraph{Latent Space Representation of Characters in Tolkien's Legendarium} 

Representation learning is a common challenge both in natural language processing and graph learning.
In NLP, word or text embedding techniques are commonly used to generate vector space representations that can then be used to apply downstream machine learning techniques to literary texts.
A popular word embedding technique is \texttt{word2vec} \cite{mikolov2013distributed,mikolov2013efficient}, which we use as a baseline in our analysis that only uses the text corpus while being agnostic to the topology of the character network.
We specifically use the \texttt{SkipGram} architecture to train a neural network with a single hidden layer with $d$ neurons that captures the context of words in the text corpus, i.e. a concatenation of all works in our corpus.
The weights of neurons in the hidden layer are then interpreted as positions of words in a $d$-dimensional latent vector space.
For our analysis, we used the \texttt{word2vec} implementation in the package \texttt{gensim} with default parameters, i.e. we use a latent space with $d=300$ dimensions.

\begin{figure}[!ht]
    \centering
    \includegraphics[width=.5\textwidth]{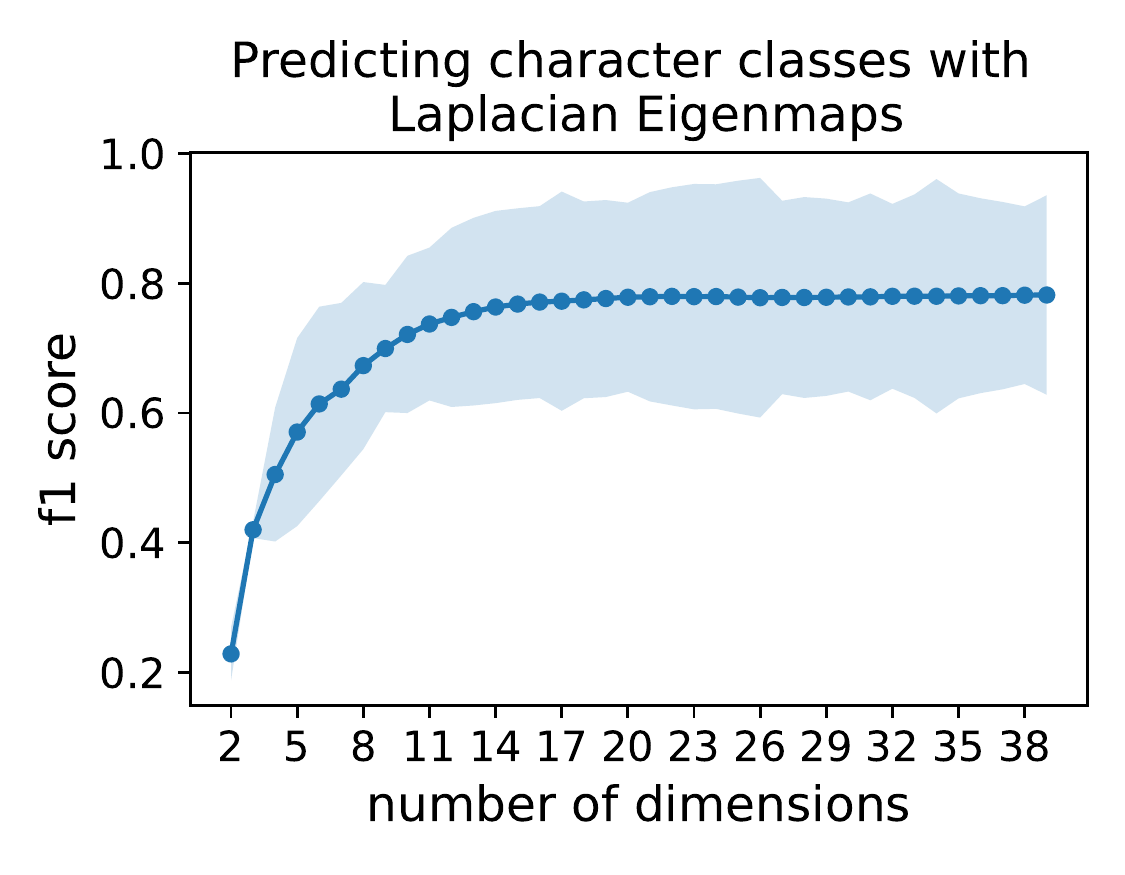}
    \caption{Performance of node classification (y-axis) for a Laplacian Eigenmap graph embedding of characters in Tolkien's Legendarium (with subsequent logistic regression) using different numbers of dimensions (x-axis).The shaded area indicates the standard deviation of the f1-score.\label{fig:LE_choose_dim_all5}}
\end{figure}

Apart from this standard NLP approach to generate latent space embeddings, we apply two graph representation learning techniques to generate character embeddings based on the topology of the character network:
The first approach, Laplacian Eigenmaps, uses eigenvectors corresponding to the leading eigenvalues of the Laplacian matrix of a graph \cite{belkin2003laplacian}. 
This can be seen as a factorization of an eigenmatrix \cite{qiu2019netsmf} that yields a representation of nodes in a $d$-dimensional latent space, where $d$ is chosen to be smaller than the number of nodes.
To determine a reasonable choice for the number of dimensions $d$, we performed an experiment in which we evaluated the average performance of a logistic regression model for node classification (see detailed description below) for different dimensions $d$ of the latent space.
The results of this experiment are shown in \Cref{fig:LE_choose_dim_all5}.
As expected, we observe tendency to underfit for a very small ($d<10$) number of dimensions, while the performance saturates as we increase $d$ beyond a ``reasonable'' value that allows to capture the topological patterns in the network. 
This analysis informed our choice of $d=20$ for the subsequent experiments.
As second approach we adopt the popular graph representation learning method \texttt{node2vec}, which applies the \texttt{SkipGram} architecture to sequences of nodes generated by a biased second-order random walk (i.e. a walk with one-step memory) in a graph \cite{grover2016node2vec}.
We again use $d=20$ hidden dimensions to make it comparable with the previous method.
We note that choosing a value of $d=20$ for the graph representation learning techniques, which is substantially smaller than the default value of $d=300$ for \texttt{word2vec}, is justified because the number of nodes in the Legendarium graph ($n=238$) is much smaller than the vocabulary used by \texttt{word2vec} ($n=18,430$).

We finally adopt two Graph Neural Networks, namely Graph Convolutional Networks (GCNs) \cite{Kipf:2016tc} and Graph Attention Networks (GATs) \cite{velikovi2017graph}.
Both of these deep graph learning techniques use a variant of neural message passing, the main difference being that GATs use a learnable attention mechanism that can place different weights on nearby nodes \cite{zhou2020graph}.
We trained both architectures to address the graph learning tasks, i.e. node classification and link prediction, outlined below, using a hidden layer with $d=20$ neurons.
We then use the hidden layer activations of both architectures to infer a representation of characters in a $20$-dimensional latent space.
To exclusively focus on the graph \emph{topology}, for our experiments we treated the network as an unweighted graph.
Additional results for experiments with weighted graphs are included in \Cref{sec:app:weighted}.

\begin{figure}[htp!]
    \centering  
\begin{subfigure}[c]{\textwidth}
\centering
    \includegraphics[width=\textwidth]{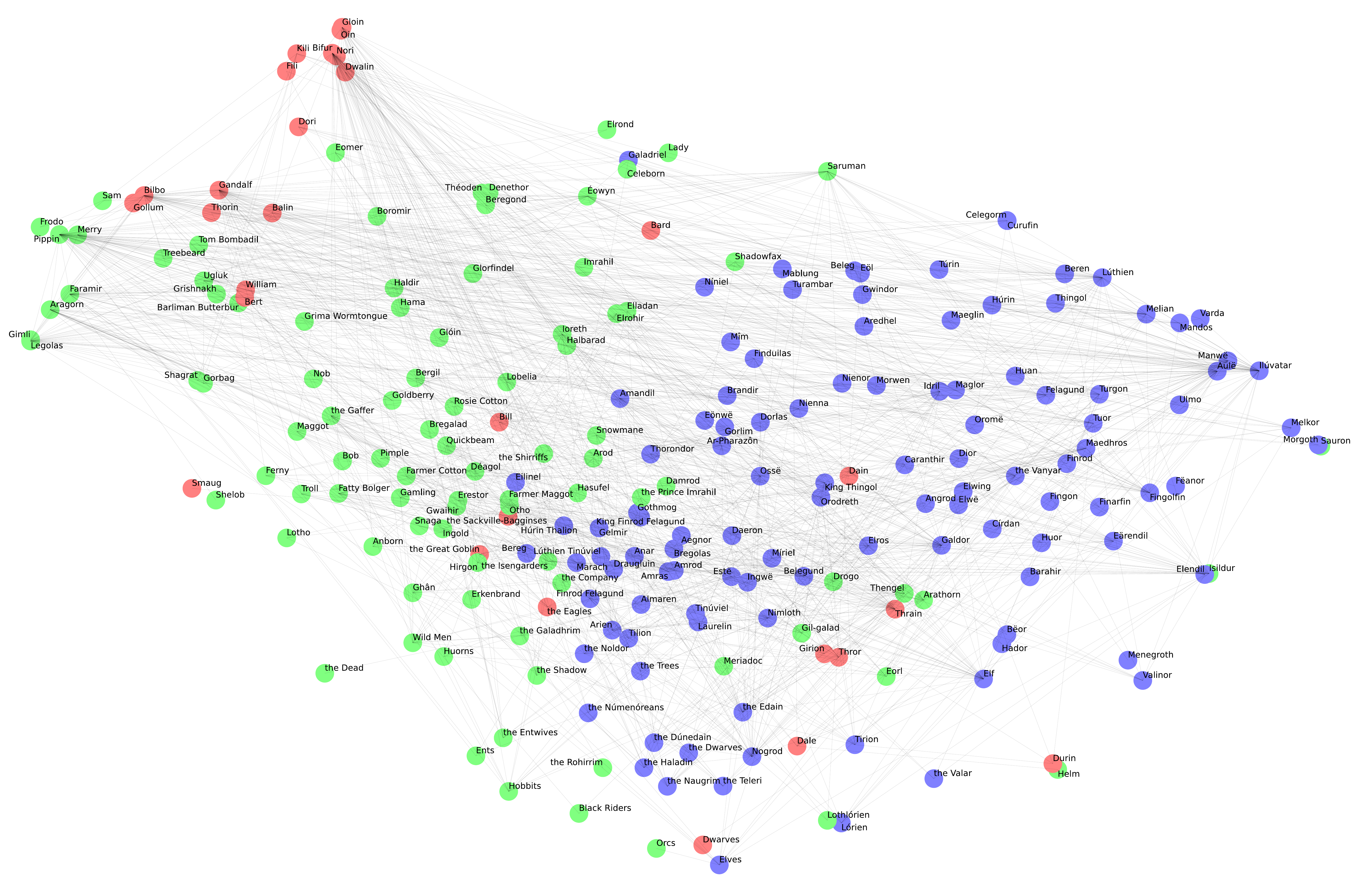}
    \caption{latent space embedding of characters obtained by applying the word embedding \texttt{word2vec} to the whole text corpus; edges represent character co-occurrences in the text}
\end{subfigure}
\begin{subfigure}[c]{\textwidth}
    \centering
    \includegraphics[width=\textwidth]{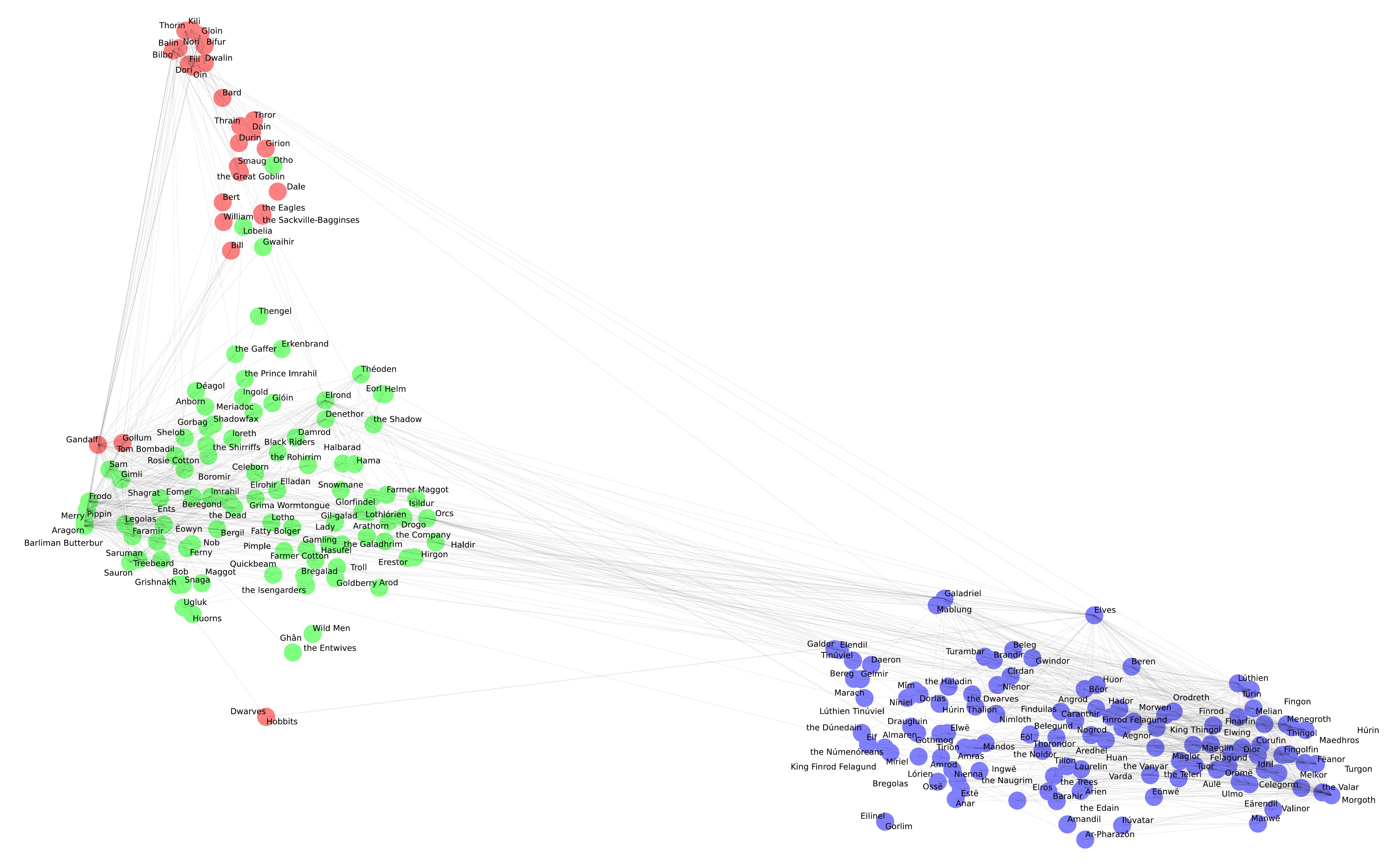}
    \caption{Graph Convolutional Network using \texttt{word2vec} character embeddings as additional node features}
\end{subfigure}
\caption{Comparison of two latent space embeddings of characters in Tolkien's Legendarium using \texttt{word2vec} (a) and the hidden layer activations of a Graph Convolutional Network, where nodes carry additional features that correspond to \texttt{word2vec} embeddings (b). Node colours indicate the work in which characters appear most frequently (red: \hobbit{}, green: \lotr{}, blue: \sil{}). Two-dimensional visualisations were generated using the dimensionality reduction technique t-SNE \cite{van2008visualizing}.\label{fig:embed:legendarium}}
\end{figure}

Leveraging the ability of GNNs to consider additional node attributes, we compare three different approaches:
First, we use GNNs without additional node features, which we emulate by initializing the message passing layer with a one-hot-encoding (OHE) of characters.
In other words, for a network with $n$ nodes, each node $i=0, \ldots, n-1$ we assign a ``dummy feature'' vector $f_i \in \mathbb{R}^{n}$ defined as: 
\[f_i=(\underbrace{0,\ldots, 0}_{i \text{ times}}, 1, \underbrace{0, \ldots, 0}_{n-i-1 \text{ times}}) \]
Second, we use the node embeddings generated by \texttt{node2vec} as additional node features that are used in the message passing layers. 
Third, we assign the \texttt{word2vec} as additional node features thus combining NLP and graph neural networks.
In \Cref{fig:embed:legendarium} we illustrate two latent space representations of characters generated by (a) \texttt{word2vec} and (b) the combination of GCN with \texttt{word2vec} features.
For both figures, we used t-SNE \cite{van2008visualizing} to reduce the latent space embedding to two dimensions, nodes are coloured according to the works in which the corresponding characters occur most frequently.
A comparison of the two embeddings clearly highlights the advantage of graph learning over a mere application of \texttt{word2vec}: Different from the word embedding, the combination of GCN with \texttt{word2vec} generates a latent space representation that captures the distinction of characters across different works in Tolkien's Legendarium.
We argue that this visualisation highlights the additional information that graph neural networks can leverage from the topology of character networks, as opposed to mere word-context pair statistics.
In the following, we more thoroughly investigate this interesting aspect in two clearly defined learning tasks.

\paragraph{Predicting Character Classes}  

We use the methods outlined above to address a supervised node classification problem in the \emph{Legendarium Graph}, i.e. the single character network capturing all works in Tolkien's Legendarium.
We assign three labels to nodes that correspond to the work (i.e. \sil{}, \hobbit{}, and \lotr{}), in which the corresponding character is most prominent.
We extract these labels automatically as $\argmax_{book} \nicefrac{\ct(c,book)}{\sum_{c \in C} \ct(c, book)}$, where $\ct(c, book)$ is the number of mentions of character $c$ in $book$ and $C$ are all characters. 
We verify the labels manually, finding them to be reasonable in all cases.
The resulting three classes contain 113 (\sil{}), 30 (\hobbit{}) and 99 (\lotr{}) characters, i.e. there is class imbalance that we address by using the macro-averaged f1-score, precision and recall.
We highlight that the information on the different works is withheld from all methods, i.e. for the word embedding and character network construction we concatenate all works to a single text corpus.
We train our models on a training set of labelled characters and use them to predict the unknown classes of unlabelled characters in a test data set.
As a baseline that does not utilise the topology of the character network, we first train a logistic regression model using the embeddings generated by \texttt{word2vec}. 
Similarly, we train a logistic regression model on the embeddings generated by Laplacian Eigenmaps and \texttt{nodev2ec}.
For \texttt{node2vec} we use three different sets of hyper-parameters $p$ and $w$, where for $p=q=1$ \texttt{node2vec} is equivalent to the graph embedding technique DeepWalk \cite{perozzi2014deepwalk}.
We trained the model for $200$ epochs.
We finally train the GCN and GAT model either using one-hot encoding (OHE) or assigning additional node features generated by the word and graph embeddings as explained above.
For both we used two message passing layers and we trained the models for $5000$ epochs using an Adam optimizer with learning rate $lr=0.0001$.
We evaluate all models using a $10-$fold cross-validation. 
Average results with standard deviation are shown in \Cref{table:nodeclassification:one}.

\begin{table}[ht!]
    \centering
    \begin{tabular}{l l l l | l}
        \toprule 
        ~ & \multicolumn{3}{c|}{\bfseries Character classification} & {\bfseries Link Prediction} \\
        ~ & F1-score & Precision & Recall & ROC/AUC \\         
        \midrule 
        word2vec\textsuperscript{300}             & 79.45 ± 0.12 & 80.67 ± 0.13 & 80.08 ± 0.12 & 78.16 ± 0.79\\ 
        \hline
        Laplacian Eigenmap\textsuperscript{20} (LE) & 81.52 ± 0.14 & 86.59 ± 0.09 & 84.13 ± 0.14 & 64.03 ± 2.28 \\ 
        \hline
        node2vec\textsuperscript{20} \textsubscript{$p=1, q=4$} & 82.10 ± 0.12 & 84.54 ± 0.12 & 84.85 ± 0.11 &  79.06 ± 0.01 \\ 
        node2vec\textsuperscript{20}\textsubscript{$p=4, q=1$} & 81.75 ± 0.12 & 81.69 ± 0.12 & 84.51 ± 0.12&  80.71 ± 0.01\\ 
        node2vec\textsuperscript{20}\textsubscript{$p=1, q=1$} & 87.07 ± 0.13 & 89.10 ± 0.12 & 87.30 ± 0.13 &  80.15 ± 0.02\\ 
        \hline 
        GCN\textsuperscript{20}\textsubscript{LE} & 71.89 ± 0.20 & 72.40 ± 0.22 & 73.24 ± 0.18 & \textbf{88.53 ± 1.25} \\ 
        GCN\textsuperscript{20}\textsubscript{node2vec\textsubscript{$p=1, q=1$}}  & 92.17 ± 0.10 & \textbf{96.69 ± 0.04} & 91.08 ± 0.11 & 88.03 ± 1.43 \\
        GCN\textsuperscript{20}\textsubscript{OHE}  & 90.55 ± 0.13 & 92.71 ± 0.12 & 89.95 ± 0.13 & 80.78 ± 1.16 \\ 
        GCN\textsuperscript{20}\textsubscript{word2vec} & \textbf{92.32 ± 0.10 }& 95.27 ± 0.08 & \textbf{91.42 ± 0.10} & 81.67 ± 1.43 \\ 
        \hline
        GAT\textsuperscript{20}\textsubscript{LE} & 53.60 ± 0.08 & 51.72 ± 0.05 & 57.78 ± 0.08 & 79.41 ± 1.83 \\
        GAT\textsuperscript{20}\textsubscript{node2vec\textsubscript{$p=1, q=1$}} & 91.18 ± 0.09  & 95.57 ± 0.04 & 90.37 ± 0.10 & 86.23  ± 0.99 \\
        GAT\textsuperscript{20}\textsubscript{OHE} & 83.52 ± 0.14 & 85.59 ± 0.15 & 84.68 ± 0.14 & 80.52 ± 1.76 \\ 
        GAT\textsuperscript{20}\textsubscript{word2vec} & 89.41 ± 0.07 & 93.98 ± 0.08 & 89.13 ± 0.10 & 82.88 ± 2.08 \\
        \bottomrule
    \end{tabular}
    \caption{Performance of word embedding (\texttt{word2vec}), graph representation learning (\texttt{Laplacian Eigenmap} and \texttt{node2vec}), and graph neural networks (GCN and GAT) in supervised character classification and link prediction for a character network capturing Tolkien's Legendarium. For the word and graph embedding techniques, a logistic regression model was used for classification and link prediction. \label{table:nodeclassification:one}}
\end{table}

Interestingly, we find that the performance of character classification based on the word embedding \texttt{word2vec} is worse than \emph{any} of the graph learning techniques, thus highlighting the added value of the graph perspective.
The graph embedding technique \texttt{node2vec}, which uses a SkipGram architecture to embed nodes, clearly outperforms the graph-agnostic \texttt{word2vec}, despite both using the same logistic regression model for the class prediction.
Moreover, we find that a simple application of Laplacian Eigenmaps, i.e. a mathematically principled matrix decomposition, yields comparable node classification performance than the neural network-based \texttt{node2vec} embedding.
The best precision is achieved for a Graph Convolutional Network (GCN) with additional node features generated by \texttt{node2vec}, while the best f1-score and recall are achieved for GCN with additional \texttt{word2vec} embeddings.
We attribute this to the fact that the combination of word embeddings and graph neural networks integrates two complementary sources of information, thus highlighting the advantages of methods that leverage both NLP and graph learning.

\begin{figure}[htp!]
    \centering 
\begin{subfigure}[c]{\textwidth}
\centering
    \includegraphics[width=\textwidth]{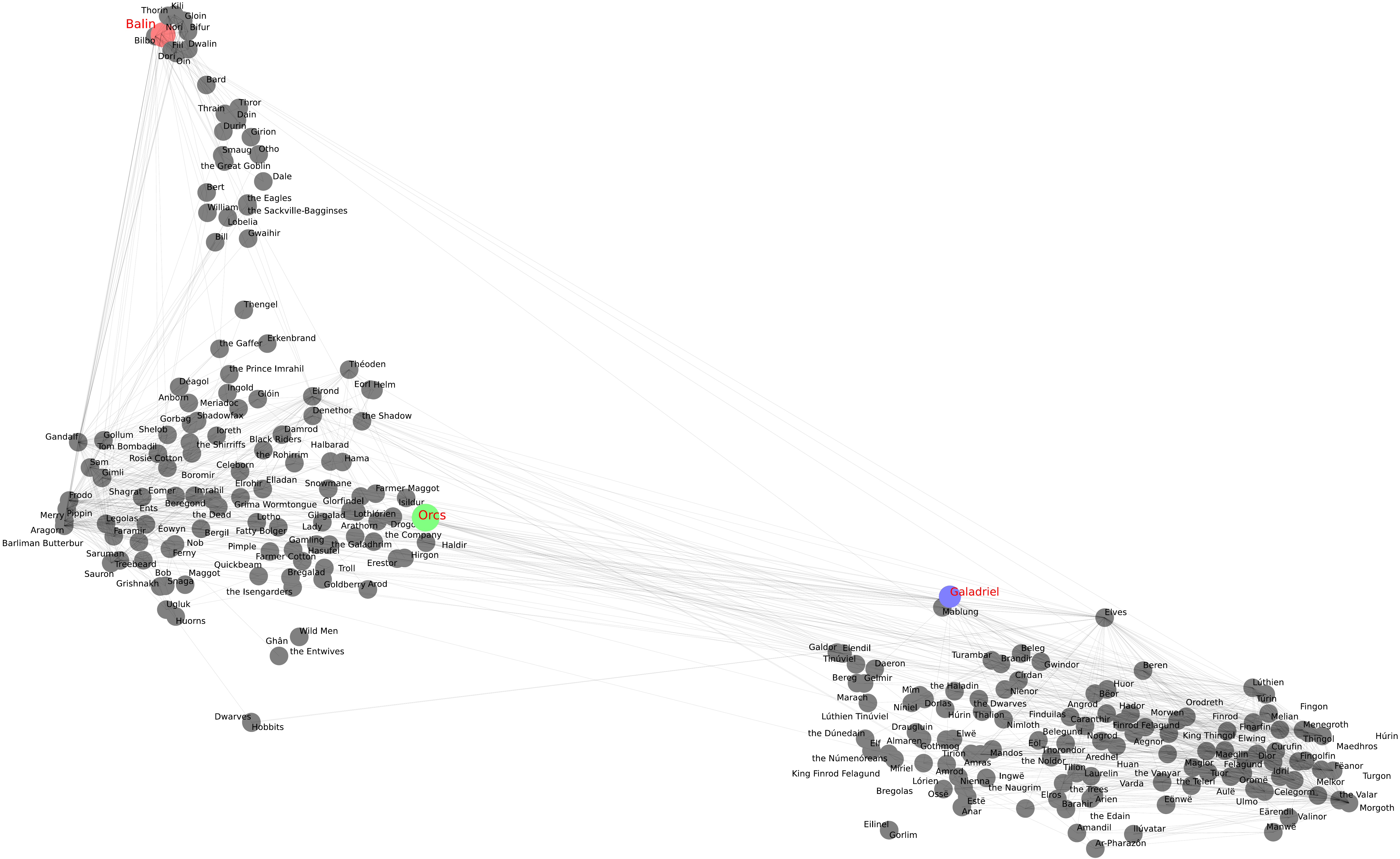} 
    \caption{Training network with three labelled characters \emph{Galadriel}, \emph{Orcs}, and \emph{Balin} (shown as coloured nodes).}
\end{subfigure}
\begin{subfigure}[c]{\textwidth} 
    \centering
    \includegraphics[width=\textwidth]{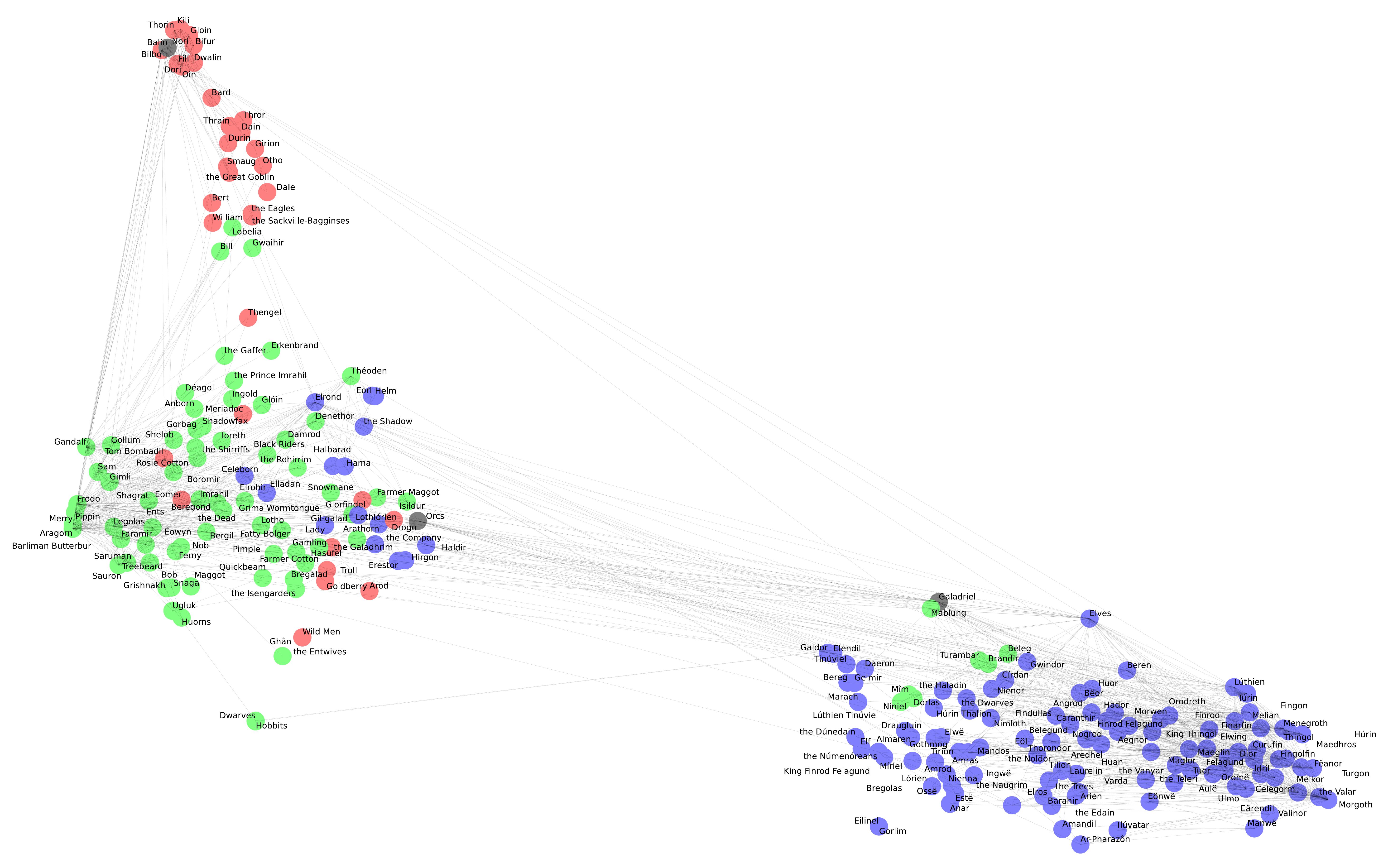}
    \caption{Character classes predicted by Graph Convolutional Network (GCN) using \texttt{word2vec} character embeddings as additional node features.}
\end{subfigure}
\caption{Illustration of semi-supervised character classification, where the training set contains only three randomly chosen labelled characters. Node colours indicate ground truth classes (a) and class predictions by Graph Convolutional Network (GCN) (b) (red: \hobbit{}, green: \lotr{}, blue: \sil{}). Two-dimensional visualisations were generated using the dimensionality reduction technique t-SNE \cite{van2008visualizing}. Node positions correspond to \Cref{fig:embed:legendarium} (b), which can be used for a comparison with ground truth character classes. Remarkably, a GCN model with additional \texttt{word2vec} features achieves an f1-score of $\approx 79.7 \%$ with an associated precision of $ \approx 78.6 \%$ and recall of $\approx 82.6 \%$. \label{fig:semi}}
\end{figure}

In \Cref{fig:semi} we further demonstrate the ability of Graph Neural Networks to perform a semi-supervised classification, i.e. their ability to accurately predict classes based on a very small number of labelled examples.
\Cref{fig:semi}(a) shows the training network with three randomly coloured characters, one for each ground-truth class. 
Nodes in the test set are shown in grey and node positions are chosen based on the latent space representation shown in \Cref{fig:embed:legendarium}(b).
We use these three labelled characters to train a GCN with additional \texttt{word2vec} embeddings as node features.
We then use the trained model to predict the character classes in the test set.
The predicted classes are shown as coloured nodes in \Cref{fig:semi}(b), where the three training nodes are shown in grey.
A visual comparison of \Cref{fig:semi}(b) and \ref{fig:embed:legendarium} (b) allows to evaluate the prediction against the ground truth. 
Despite the very sparse labelled examples, and thanks to its use of the graph topology of the unlabelled nodes in the training set, the GCN model is able to accurately predict character classes, reaching an f1-score of $\approx 79.7 \%$, a precision of $ \approx 78.6 \%$ and a recall of $\approx 82.6 \%$.
Remarkably, this shows that the combination of a GCN model with \texttt{word2vec} node features yields a higher f1-score and recall with only three labelled examples than a word embedding alone, even when all characters in the training set are labelled.

\paragraph{Predicting Character Interactions} 

We finally address link prediction, which refers to the task of predicting ``missing'' links in a graph, i.e. links that are either ``missing'' in incomplete data or that likely form in the future.
Link prediction is a well-studied graph learning problem, with important applications in social network analysis and recommender systems \cite{lu2011link}.
In the context of character networks, it is relevant because it could be used to alleviate the low recall of the rigid, sentence-based character network extraction that we employed in \Cref{sec:network_analysis}.

Adopting a supervised approach, we split the edges of the \emph{Legendarium Graph} in a training and set set, where we withhold 10 $\%$ of the edges during training to test our model.
We use \texttt{word2vec}, Laplacian Eigenmaps and \texttt{node2vec} to generate embeddings $f_v \in \mathbb{R}^d$ of characters.
For \texttt{word2vec} embeddings are generated using the full text corpus. 
For the graph embedding techniques Laplacian Eigenmaps and \texttt{node2vec} we only use links in the training set, potentially putting them at a disadvantage in terms of training data.
We use the resulting feature vectors to calculate the element-wise (Hadamard) product $f_v \circ f_w \in \mathbb{R}^d$ for character pairs $v, w$, which yields $d$-dimensional features for all character pairs.
We then use features of positive instances (i.e. pairs $v, w$ connected by a link $(v,w)$ in the training set) and negative instances (pairs $v,w$ not connected by a link in the training set) to train a (binary) logistic regression classifier and use the trained model to predict links in the test set.
We use negative sampling to mitigate the imbalance between negative and positive instances in the training set.
For the two GNN architectures we adopt the common approach to add a \emph{decoding} step that computes the Hadamard product of node features after the last message passing layer.
We use a Binary Cross Entropy with Logits Loss function and train the models for $15000$ epochs using an Adam Optimizer with learning rate $0.001$.

We use the Receiver-Operating Characteristic (ROC) to evaluate our models, i.e. we compute ROC curves that give the true and false positive rates across all discrimination thresholds.
An example (and explanation) is included in \Cref{sec:app:ROC}.
We compute the Area Under Curve (AUC) of ROC curves within the unit square, which range from $0$ to $1$. 
$0.5$ corresponds to the performance of a random classifier, values $<0.5$ indicate worse and values $>0.5$ better than random performance.
We again evaluate all models using a $10-$fold cross-validation. 
\Cref{table:nodeclassification:one} reports average results and standard deviations of the AUC for all models. 
With the exception of Laplacian Eigenmaps, we find that graph methods generally perform better than \texttt{word2vec}.
We further observe that GNNs perform considerably better than \texttt{node2vec}, where the best performance is achieved when coupling GCN with Laplacian Eigenmaps or \texttt{node2vec}.

\section{Conclusion}
\label{sec:conclusion}

In summary, we used natural language processing techniques like named entity recognition and coreference resolution to construct a single character network from a corpus of works that constitute J.R.R. Tolkien's Legendarium.
Apart from characterising the network based on social network analysis, we adopt state-of-the-art graph learning techniques to (i) generate latent space embeddings of characters, (ii) automatically classify characters based on the work to which they belong, and (iii) predict character co-occurrences.
For all three tasks, we find a significant advantage of Graph Neural Networks (GNNs) over a common word embedding technique and we find that a combination of both yields the best performance.
Our approach to construct a single graph for multiple literary texts could be interesting to analyze other corpora of works with overlapping characters (e.g. mythology, historical novels, etc.). 
We further believe that our results on the application of GNNs to address a link prediction task have interesting implications for computational literary studies.
Considering the difficulty of coreference resolution, and the low coverage of the resulting character networks that we observed in our experiments, we expect that link prediction could potentially be used as an approach to address the low recall observed for the sentence-based co-occurrence networks.


\bibliography{bibliography,albins}

\newpage
\appendix
\title{Appendix}

\section{Alternative Approaches to Detect Character Co-Occurrences}
\label{sec:app:cooc}

As an alternative to the sentence-based co-occurrence approach described in \Cref{sec:data_processing}, we also evaluated two other approaches in initial experiments: detection based on (i) the parse tree and (ii) a sliding text window.
For (i), we marked all sentences as an ``interaction'' between two characters if they had the same head word in the parse tree.
This is the most strict version of our character network construction, as it only captures explicit interactions (e.g., ``Frodo saw Sam'').
Due to the small number of detected interactions, we discarded this strategy.
On the other hand, (ii) is more lenient than our final sentence-based approach, only requiring two characters to be mentioned within a window of a fixed number of characters (letters). 
This approach of extracting character co-occurrences within a sliding text window has been adopted in a number of prior works \cite{beveridge2016network,bonato2016mining}.
It allows to capture interactions between characters that are not mentioned in the same sentence, but introduces the risk of detecting a large number of spurious interactions.
In our experiments, we chose a window size of 2000 characters, with the additional restriction that chapter borders may not be crossed.
Aiming for character networks that maintain a balance between recall (i.e. detecting all meaningful character links) and precision (i.e. limiting the number of spurious links), we decided to use the sentence-based interaction detection.

\section{Narrative Charts for \hobbit{} and \sil{}}
\label{sec:app:narrative}

\begin{figure}[!ht]
    \includegraphics[width=\textwidth]{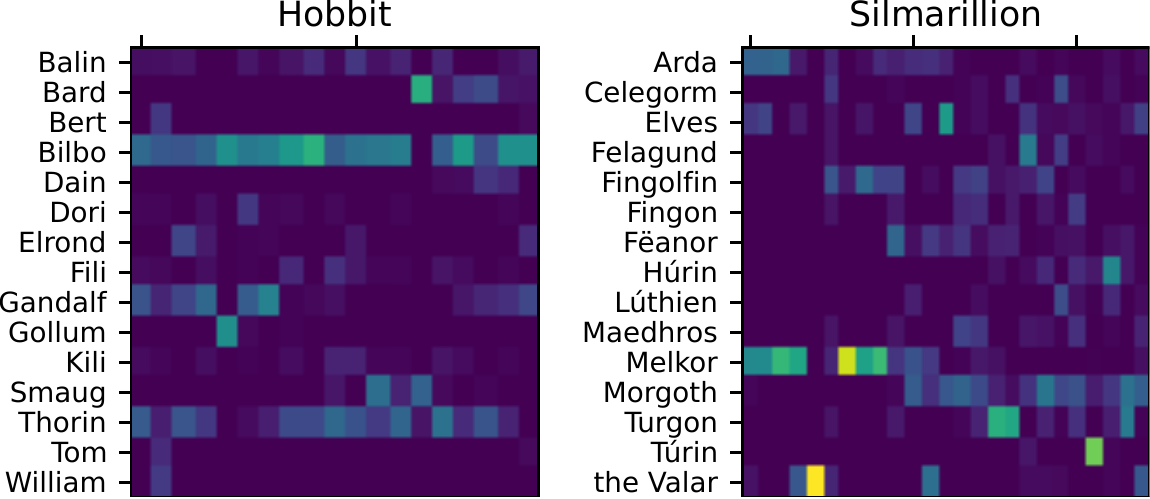}
    \caption{Narrative charts for \hobbit{} (left) and \sil{} (right)\label{fig:app:narrative}}
\end{figure}

Complementing the results in \Cref{sec:data_processing}, in \Cref{fig:app:narrative} we present two additional narrative charts that we generated for \hobbit{} and \sil{}.

\pagebreak

\section{Visualisation of Additional Latent Space Embeddings}
\label{sec:app:embeddding}

\begin{figure}[!htp]
     \centering
     \begin{subfigure}[c]{\textwidth}
        \centering
         \includegraphics[width=\textwidth]{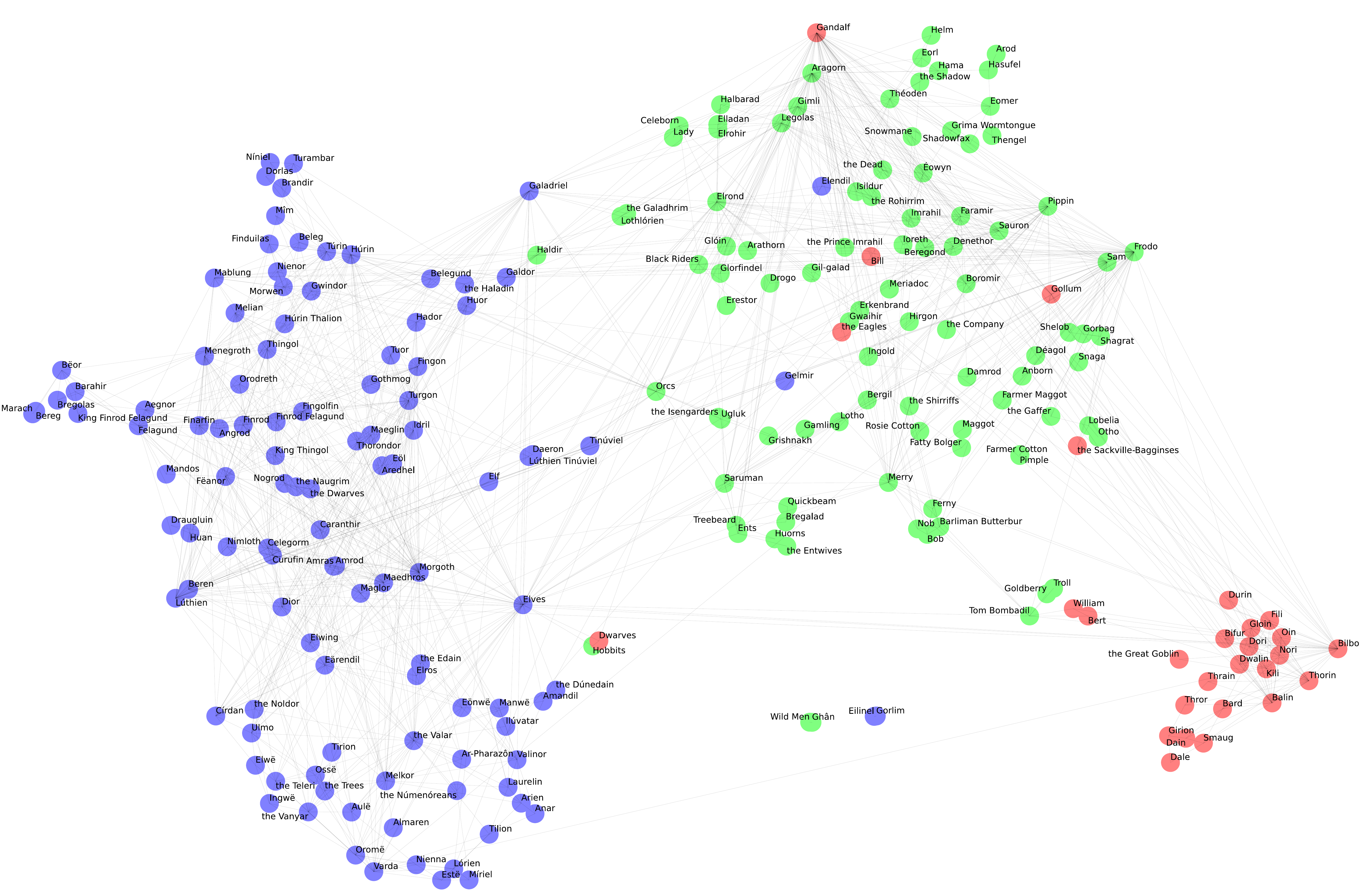}         
         \caption{\texttt{node2vec}}
     \end{subfigure}
     \begin{subfigure}[c]{\textwidth}
        \centering
         \includegraphics[width=\textwidth]{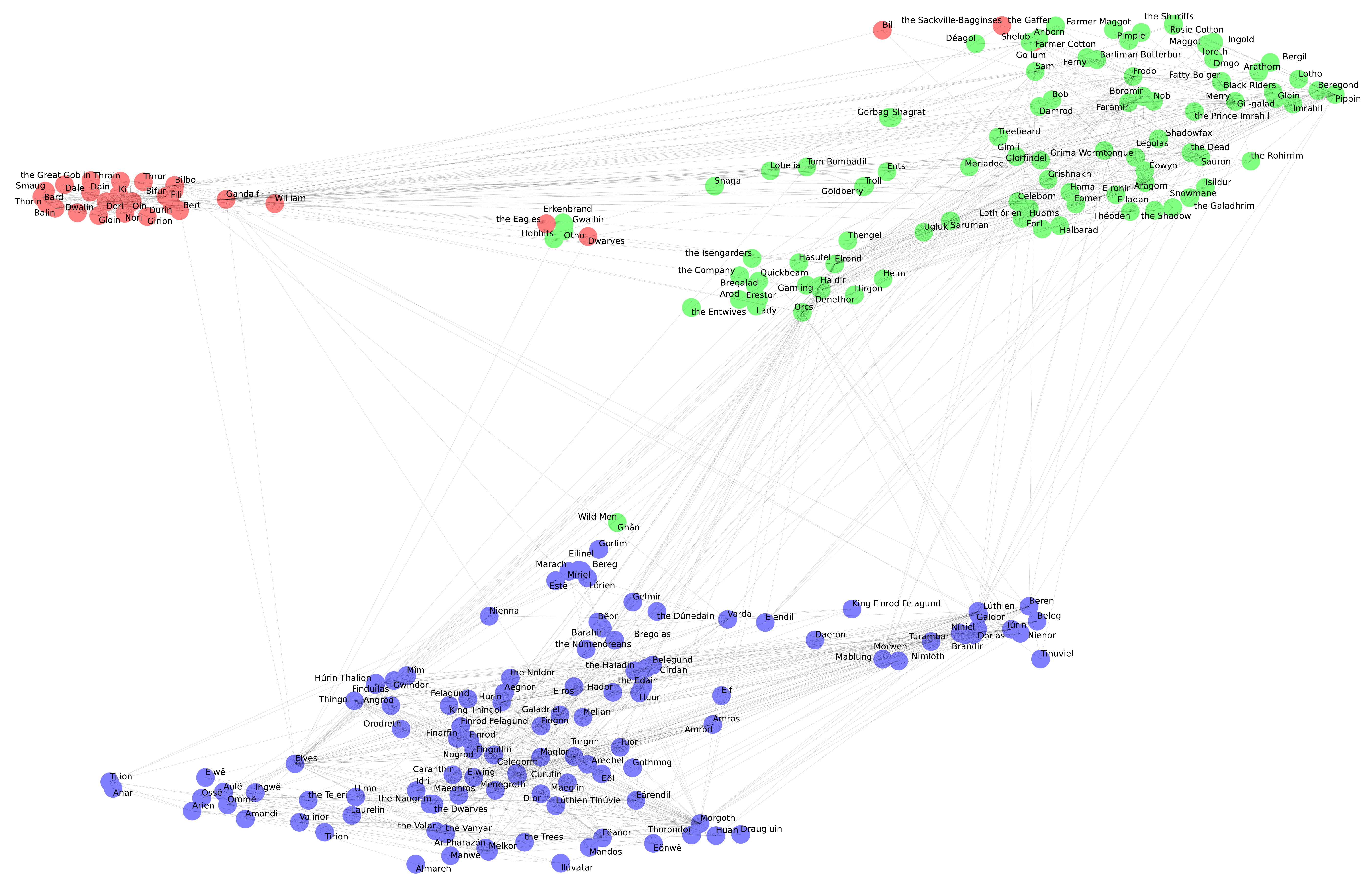}
         \caption{Graph Attention Networks with additional \texttt{word2vec} features}
     \end{subfigure}
    \caption{Latent space embedding of characters in Tolkien's Legendarium using \texttt{node2vec} (a) and Graph Attention Networks with additional \texttt{node2vec} features (b).\label{fig:app:embed:1}}
\end{figure}

\begin{figure}[!htp]
     \centering 
     \begin{subfigure}[c]{\textwidth}
        \centering
         \includegraphics[width=\textwidth]{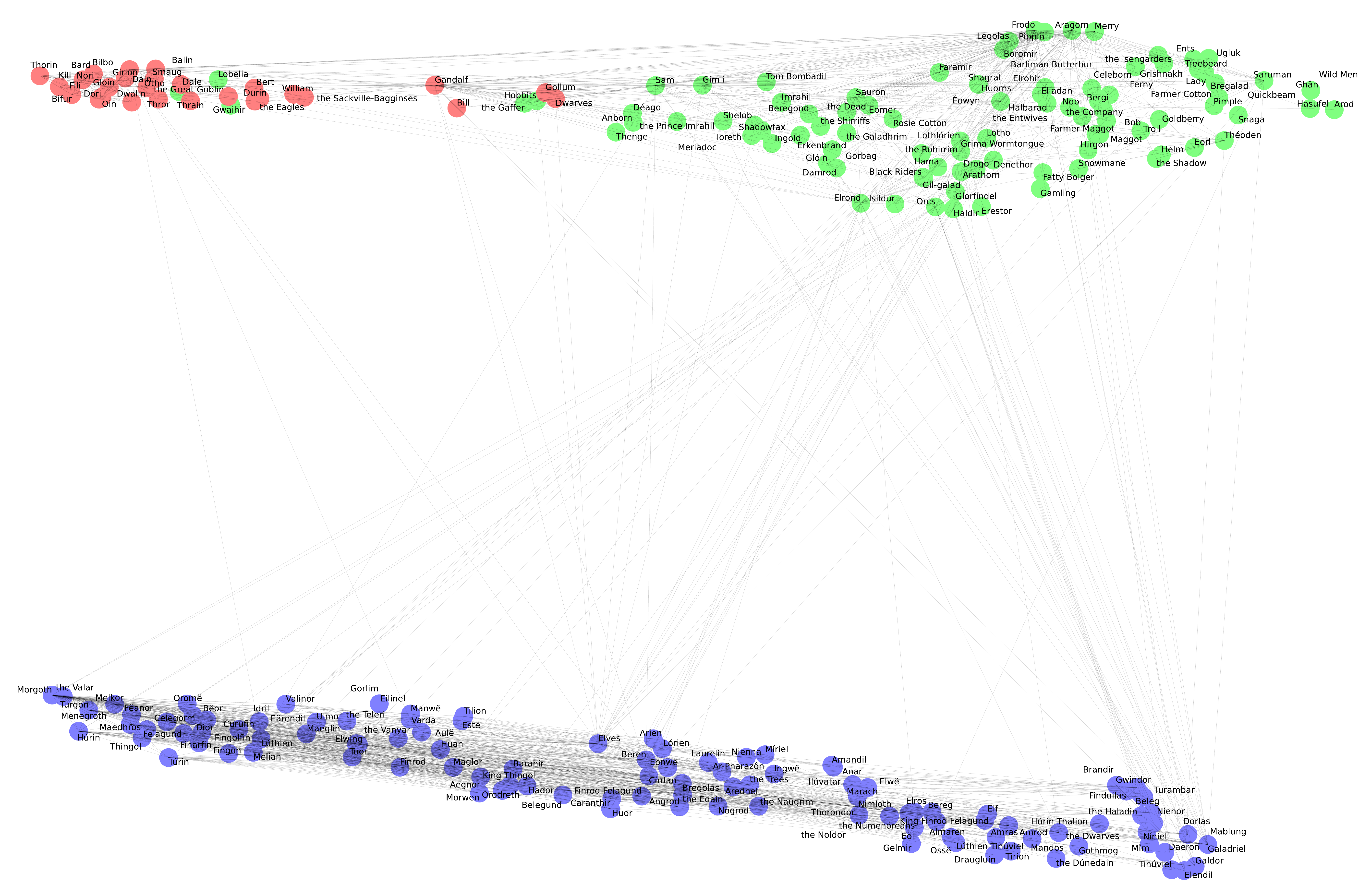}
         \caption{Graph Convolutional Networks with one-hot-encoding (OHE)}
     \end{subfigure} 
     \begin{subfigure}[c]{\textwidth}
        \centering
         \includegraphics[width=\textwidth]{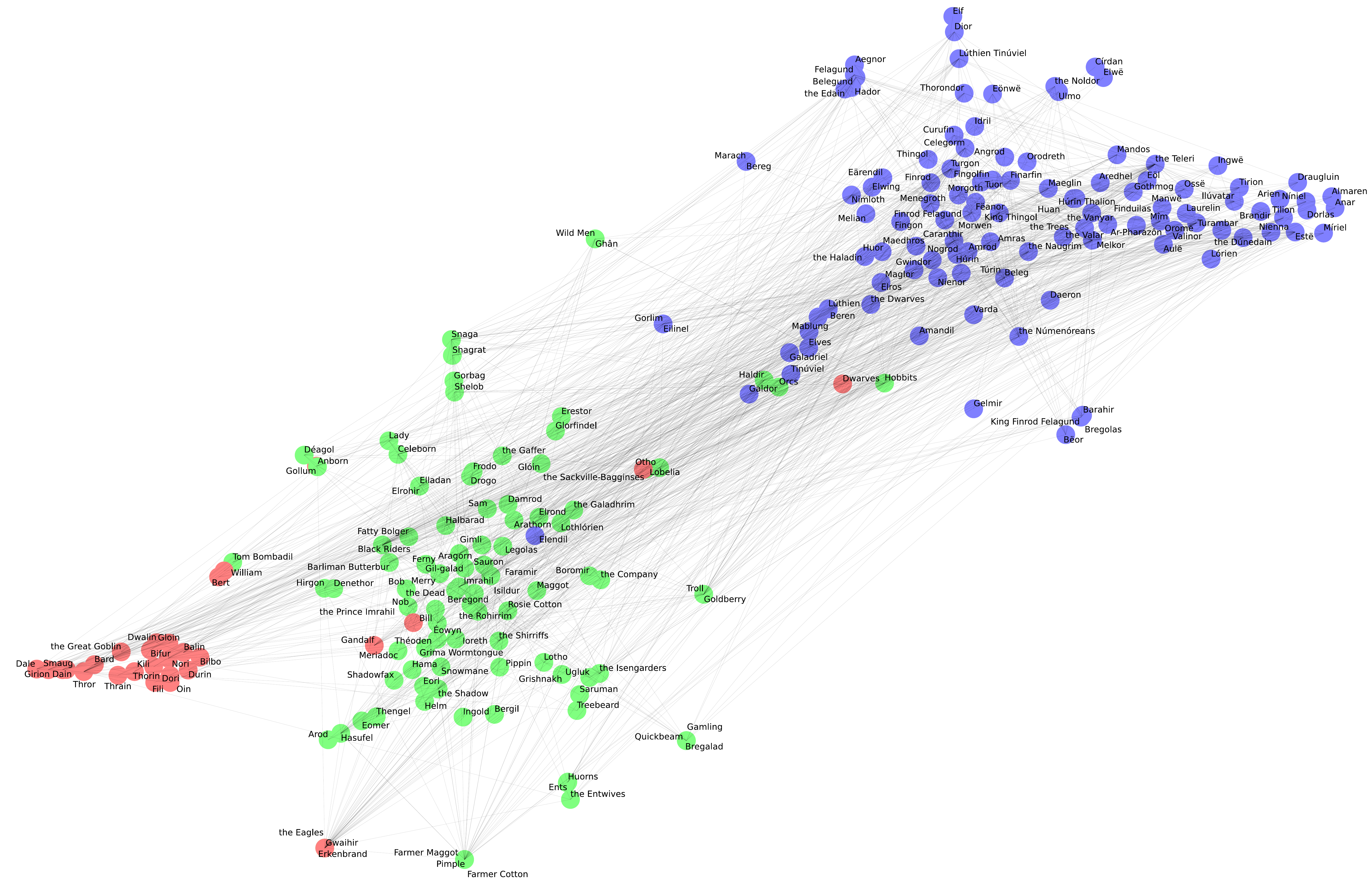}   
         \caption{Laplacian Eigenmap}
     \end{subfigure}
    \caption{Latent space embedding of characters in Tolkien's Legendarium using GCN with one-hot encoding (OHE) (a) and Laplacian Eigenmap (b).\label{fig:app:embed:2}}
\end{figure}

In \Cref{fig:app:embed:1} and \Cref{fig:app:embed:2} we include additional visual representations of latent space embeddings of characters obtained by those methods that have not been included in the main text.
For all methods, we employed t-SNE \cite{van2008visualizing} to reduce the dimensionality of the latent space embeddings to two dimensions.

\section{Node Classification and Link Prediction with Weighted GNNs} 
\label{sec:app:weighted}

Complementing the results discussed in the main article, in \Cref{tab:app:weighted} we include additional results for which we applied Graph Neural Networks to \emph{weighted} graphs, i.e. different from \Cref{table:nodeclassification:one} we consider a character network with link weights $w((v,w))$ that capture the number of co-occurrences of two characters $v,w$.
Due to time constraints, we performed this analysis only for the best performing Graph Neural Network, i.e. GCN.
These additional results suggests that, at least for our corpus, the inclusion of link weights does not significantly improve the performance of models for character classification and link prediction.

\begin{table}[ht!]
    \centering
    \begin{tabular}{l l l l | l }
        \toprule 
        ~ & \multicolumn{3}{c|}{\bfseries Character classification} & {\bfseries Link Prediction} \\
        ~ & F1-score & Precision & Recall & ROC/AUC \\         
        \midrule 
        GCN\textsuperscript{20}\textsubscript{LE} & 60.85 ± 0.12 & 60.84 ± 0.16 & 63.40 ± 0.10 & 79.41 ± 1.83\\ 
        GCN\textsuperscript{20}\textsubscript{node2vec\textsubscript{$p=1, q=1$}}  & 92.14 ± 0.11 & {\bfseries 95.87 ± 0.07} & 91.28 ± 0.11 & {\bfseries 86.23 ± 0.99} \\
        GCN\textsuperscript{20}\textsubscript{OHE}  & 87.92 ± 0.10 & 93.02 ± 0.08 & 86.36 ± 0.11 & 80.52 ± 1.76 \\ 
        GCN\textsuperscript{20}\textsubscript{word2vec} & {\bfseries 92.96 ± 0.10} & 95.67 ± 0.07 & {\bfseries 92.03 ± 0.10} & 82.88 ± 2.08\\ 
        \bottomrule
    \end{tabular}
    \caption{Additional results for node classification in the \emph{Legendarium} graph where links carry weights that capture the number of character co-occurrences \label{tab:app:weighted}.}
\end{table}

\section{Explanation of ROC/AUC Evaluation of Link Prediction}
\label{sec:app:ROC}

\begin{figure}[!htp]
    \centering
    \includegraphics[width=0.5\textwidth]{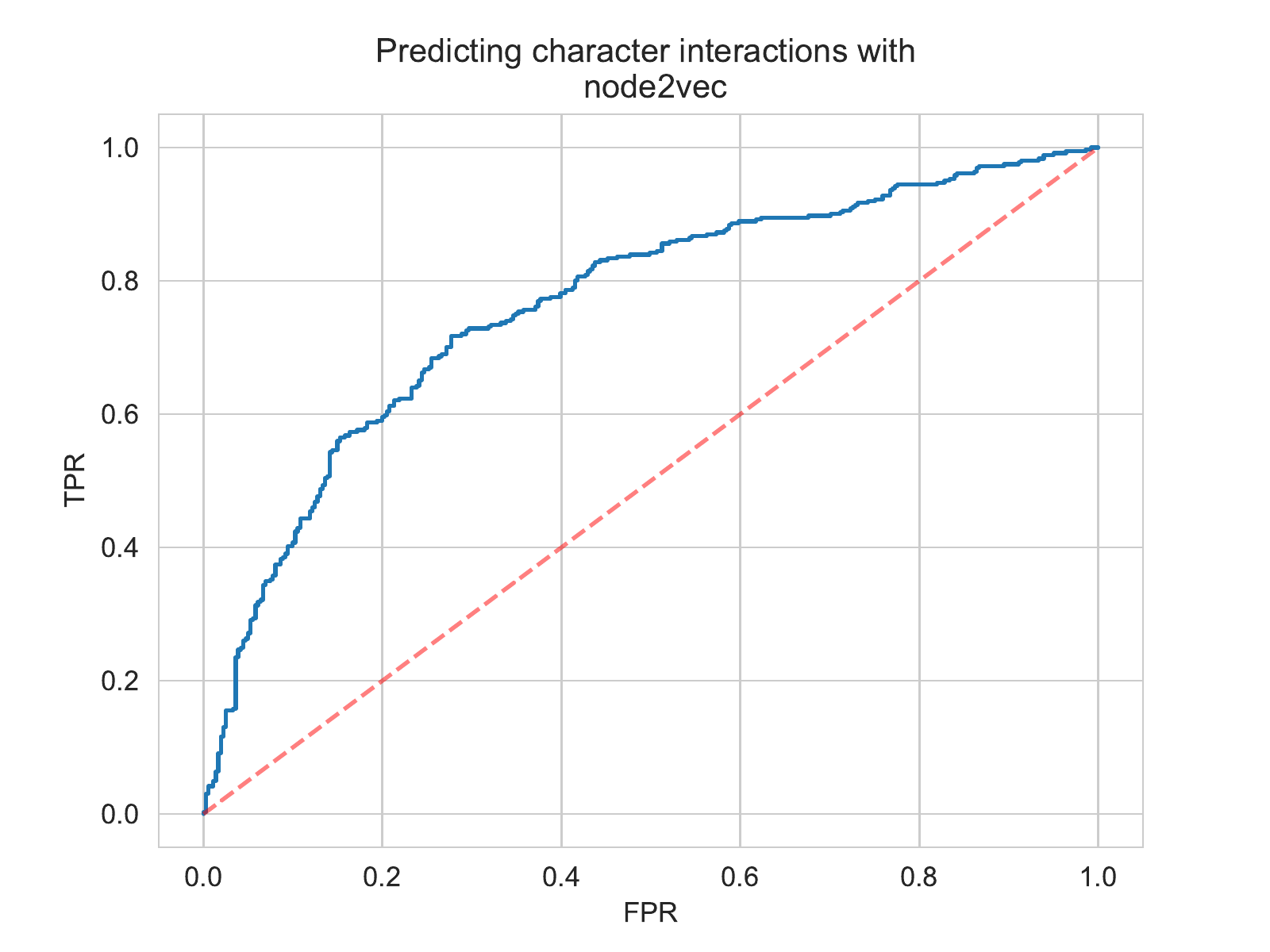}
   \caption{Exemplary ROC curve for link prediction using \texttt{node2vec} embedding and logistic regression \label{fig:app:roc}}
\end{figure}

In \Cref{sec:gnn} we use the Area Under Curve (AUC) of a Receiver-Operating Characteristic (ROC) curve to evaluate the performance of our models in link prediction, which is a common approach to evaluate the diagnostic quality of binary classifiers in information retrieval and machine learning.
A key advantage of this approach is that it enables us to evaluate the performance of a binary classifier across all possible discrimination thresholds, which can be adapted to tune the sensitivity/specificity of the prediction depending on application requirements.
To assist the reader to follow this evaluation approach, below we explain one exemplary ROC curve obtained for a link prediction in the \emph{Legendarium} graph using the a \texttt{node2vec} embedding of characters and a logistic regression model. 
To generate this curve, we first consider the prediction scores (i.e. in the case of logistic regression the positive class probability) assigned to each node pair in the test set, where a link is predicted whenever the score is above a given discrimination threshold $\epsilon$.
For each value of $\epsilon$ we can now calculate the true and false positive rate (TPR and FPR), i.e. the fraction of those predicted links for which the prediction is correct and the fraction of unconnected node pairs for which a link is predicted errorneously.
A sweep over all possible discrimination thresholds $\epsilon$ now yields a ROC curve in the unit square. 
In \Cref{fig:app:roc} we show the ROC curve of a logistic regression model using \texttt{node2vec} features.
A classifier that perfectly classifies all instances in the data will assume initial values of FPR=0 and TPR=0 only for the maximal discrimination threshold of $\epsilon=1$, where all instances are assigned to the negative class.
For any $\epsilon$ smaller than the maximum and larger than the minimum value of zero, a perfect classifier correctly predicts all instances, which yields TPR=1 and FPR=0.
For the minimum value of $\epsilon=0$, the classifier necessarily predicts the positive class for all instances, which yields TPR=1 and FPR=1.
We thus find  that the ROC curve of a perfect classifier follows the left and upper border of the unit square, which yields an Area Under Curve (AUC) of one. 
Conversely, the ROC curve of a classifier that consistently predicts the opposite of the true class follows the bottom and right border of the unit square, which yields an Area Under Curve (AUC) of zero.
For a classifier with no diagnostic ability, the FPR and TPR are expected to increase equally as we lower the discrimination threshold $\epsilon$, i.e. the ROC curve follows the so-called \emph{diagonal of no-discrimination} (see red dashed line in \Cref{fig:app:roc}) and the Area Under Curve is expected to be close to $0.5$.

\section{Evaluation of Link Prediction for Individual Works}
\label{sec:app:linkpred:single}

In the main text we present and discuss the performance of different techniques to predict links in a single character network that spans Tolkien's Legendarium.
Apart from this analysis, we additionally evaluated link prediction in character co-occurrence networks that have been generated for the three works of our corpus, i.e. \sil{}, \hobbit{}, and \lotr{} separately. 
For the sake of completeness, we include these results in \Cref{tab:app:linkpred:single} below.
Like in \Cref{sec:app:weighted}, we applied the Graph Convolutional Network (GCN) model on a weighted graph, while Graph Attention Networks (GAT) where applied to an unweighted graph\footnote{We resorted to an unweighted graph due to a supposed implementation error in the weighted GAT implementation in version 2.0.4 of the graph learning library \texttt{pytorch-geometric} \cite{fey2019fast}.}.

\begin{table}[htp!]
    \centering
    \begin{tabular}{l l l l }
        \toprule 
        ~ & \bfseries \sil{} & \bfseries \hobbit{} & \bfseries \lotr{} \\
        ~ & ROC/AUC & ROC/AUC & ROC/AUC \\         
        \midrule 
        GCN\textsuperscript{20}\textsubscript{LE} & 80.38 ± 1.97 & 71.87 ± 5.94 & 76.25 ± 2.05\\ 
        GCN\textsuperscript{20}\textsubscript{node2vec\textsubscript{$p=1, q=1$}}  & 81.74 ± 2.3 & 63.86 ± 5.57 & 76.61 ± 2.87\\ 
        GCN\textsuperscript{20}\textsubscript{OHE}  & 74.78 ± 1.9 &70.22 ± 8.76 & 63.15 ± 4.37 \\ 
        GCN\textsuperscript{20}\textsubscript{word2vec} & 76.82 ± 2.68 &70.23 ± 7.15 & 67.42 ± 2.77 \\ 
        \midrule         
        GAT\textsuperscript{20}\textsubscript{LE} & 68.99 ± 3.46 & 76.88 ± 4.53 & 61.3 ± 7.38\\ 
        GAT\textsuperscript{20}\textsubscript{node2vec\textsubscript{$p=1, q=1$}}  &79.37 ± 2.13 & 71.21 ± 6.53 & 71.92 ± 2.69 \\ 
        GAT\textsuperscript{20}\textsubscript{OHE}  & 70.88 ± 3.12& 71.52 ± 8.14 & 63.53 ± 3.68 \\ 
        GAT\textsuperscript{20}\textsubscript{word2vec} & 73.72 ± 4.13 & 72.29 ± 7.78& 64.52 ± 4.29 \\ 
        \midrule
        GCN\textsuperscript{20}\textsubscript{LE} (weighted) & 80.4 ± 1.44 & 69.84 ± 7.23 & 81.31 ± 1.96\\ 
        GCN\textsuperscript{20}\textsubscript{node2vec\textsubscript{$p=1, q=1$}} (weighted) & 82.7 ± 1.92 & 68.09 ± 7.42 & 76.46 ± 3.56 \\ 
        GCN\textsuperscript{20}\textsubscript{OHE} (weighted)  & 74.78 ± 3.3 & 67.25 ± 9.41 & 63.15 ± 3.12 \\ 
        GCN\textsuperscript{20}\textsubscript{word2vec} (weighted) & 78.8 ± 1.8 & 62.97 ± 22.35 & 64.53 ± 1.79\\ 
        \bottomrule
    \end{tabular}
    \caption{Additional results for link prediction in character co-occurrence networks constructed for individual works in our corpus. \label{tab:app:linkpred:single}.}
\end{table}

\end{document}